\documentclass[
	a4paper, % Paper size, use either a4paper or letterpaper
	10pt, % Default font size, can also use 11pt or 12pt, although this is not recommended
	unnumberedsections, % Comment to enable section numbering
	twoside, % Two side traditional mode where headers and footers change between odd and even pages, comment this option to make them fixed
]{LTJournalArticle}

\runninghead{Synthetic LiDAR Data and Downsampling on Edge Hardware} % A shortened article title to appear in the running head, leave this command empty for no running head

\title{Synthetic LiDAR Data Generation and Deterministic Downsampling \\ for Point Cloud Classification on the Edge} % Article title, use manual lines breaks (\\) to beautify the layout

 \author{
     Niclas Meyer\orcidlink{0009-0008-3039-3569}\textsuperscript{1}$^,$\thanks{ \href{mailto:niclas.meyer@s2020.tu-chemnitz.de}{niclas.meyer@s2020.tu-chemnitz.de}} 
     \hspace{0.5em}
     Stefan Reitmann\orcidlink{0000-0003-0283-8272}\textsuperscript{1}$^,$\thanks{\href{mailto:stefan.reitmann@informatik.tu-chemnitz.de}{stefan.reitmann@informatik.tu-chemnitz.de}}
}

\date{\footnotesize\textsuperscript{\textbf{1}}Chemnitz University of Technology}

\renewcommand{\maketitlehookd}{%
	\begin{abstract}
		\noindent Deploying three-dimensional deep learning frameworks to low-power embedded processors is bottlenecked by the unstructured nature of spatial data and the resource-expensive distance sorting algorithms that are often deployed before the neural networks themselves. To bridge this gap, this paper presents a hardware-constrained workflow optimized for native execution on the Raspberry Pi 5. 
        To address the reality gap of noiseless and clean computer-aided design (CAD) datasets, we utilize a physics-based simulation to construct a synthetic LiDAR repository. % BLAINDER left out here, as well as  modeled after a Velodyne UltraPuck scanner. 
        Cross-dataset evaluations demonstrate an immense drop in classification accuracy when networks optimized on clean CAD files are subjected to synthetic LiDAR sensor data -- highlighting the critical necessity of sensor-aware training. To break the latency bottleneck of traditional geometric preprocessing layers on edge CPUs, we integrate an isolated, feature-driven Critical Points Layer (CPL) as a  frontend filter. Our results show that the pre-trained CPL successfully compresses raw 1024-point clouds deterministically down to a subset of 40 to 60 unique coordinates.
        When profiled on the ARM Cortex-A76 processor, the pipeline achieves an inference throughput of approximately 50 FPS while maintaining a high instance classification accuracy of $88.36\%$, demonstrating the real-time viability of deterministic 3D perception at the edge.
	\end{abstract}
}

\begin{document}

\maketitle % Output the title section

%----------------------------------------------------------------------------------------
%	ARTICLE CONTENTS
%----------------------------------------------------------------------------------------

\section{Introduction}

Advancements in autonomous driving and robotic navigation fundamentally rely on the precise capture and real-time processing of environmental distance data \cite{li_lidar_2020}. 
To achieve this, Light Detection and Ranging (LiDAR) sensors have emerged as an industry standard. These sensors capture their surroundings as dense, unstructured 3D point clouds.  
A point cloud is a collection of unordered 3D points in space that capture the surfaces of objects and nearby geometry. When captured by LiDAR scanners, these $(x,y,z)$ values are often accompanied by their corresponding intensity values. This renders point clouds distinct from images, as there the rigid grid yields a structured format.
%\todo{Should I say here how LiDAR data is captured?}
Not only does the domain of autonomous driving take advantage of the high spatial fidelity of LiDAR data, but fields such as 
% digital archaeology and environmental engineering also rely on it for 
cultural heritage preservation and the generation of high-resolution digital terrain models \cite{jia_ai-powered_2025}.
While physical LiDAR sensors, like the Velodyne scanners, which are used in the KITTI dataset \cite{geiger_are_2012, geiger_vision_2013}, capture the scenes in their grand nature by generating tens of thousands of points per second, the resulting data is sparse, often noisy, and prone to range-dependent jitter. In stark contrast, synthetic datasets such as ModelNet \cite{wu_3d_2015} are compiled from various CAD models and in-game footage of video games, offering a more uniform representation of the data.

The processing, such as classification and segmentation of the rich point cloud data, requires deep neural networks to achieve satisfying results (see Section \ref{sec:nn_paradigms_3d_understanding}). State-of-the-art networks such as Point Transformer have millions of parameters and are thus compute-intensive \cite{wu_point_2024}. Workstation-grade hardware, especially GPUs, is required to run these sophisticated models. When deploying such networks on resource-constrained edge hardware like the Raspberry Pi 5, inference may not be feasible as data throughput exceeds memory bandwidth and processing throughput.  
Consequently, executing 3D classification tasks under high-throughput requirements demands strict optimisation techniques and strategies that can narrow down the incoming data stream to the geometrically critical features without losing important information in the process.

This paper introduces and evaluates a staged workflow, designed and executed for efficient execution of classification algorithms of 3D point clouds on existing edge hardware.
First, in order to address the domain gap between synthetic and real-world point cloud datasets, we generate realistic LiDAR data fused with sensor noise. 
We leverage the already existing ModelNet dataset and use the Blender add-on BLAINDER \cite{reitmann_blainderblender_2021-1} for generation. 
Second, we integrate the feature-driven deterministic downsampling method, Critical Point Layer (CPL) \cite{nezhadarya_adaptive_2020}, directly into the PointNet architecture \cite{qi_pointnet_2017}.
This actively prunes the number of points in a point cloud instance and leaves only the most important points for classification behind. 
Finally, we demonstrate the low-latency execution and high-throughput performance of our method by deploying and benchmarking it on a Raspberry Pi 5.

The remainder of this paper is structured as follows: Chapter \ref{sec:related_works} presents a comprehensive literature review of existing solutions for point cloud processing, focusing specifically on point cloud classification architectures. This review section serves as a basis to examine different datasets, spatial data augmentation techniques, and common downsampling methodologies. In Chapter \ref{sec:methodology}, we leverage these theoretical insights to construct our proposed workflow methodology. There, we will explain in detail the generation of synthetic LiDAR data via BLAINDER and the integration of PointNet with the Critical Points Layer to achieve efficient data compression. Chapter \ref{sec:results} puts the resulting neural networks to the test on resource-constrained edge hardware, delivering quantitative performance and latency results. Finally, Chapter \ref{sec:conclusion} provides a conclusion of our findings and outlines potential avenues for future work.
\section{Related Works}
\label{sec:related_works}

\subsection{Point Cloud Data}

Before comparing different methods of point cloud processing and sampling, it is essential to first mark out a clear definition of a point cloud. A point cloud is a set of vectors $P = \{P_i | i = 1, \ldots, N\} \in \mathbb{R}^{N \times D}$, where $N$ is the number of points in the point cloud, and $D$ is the dimensionality (i.e., $D = 3$ if the point cloud contains only $(x,y,z)$ coordinates), representing surfaces as a sparse, unstructured sets of coordinates \cite{tan_visualizing_2023}. Therefore, finding the point with the minimal distance to another is demanding, and still an ongoing research topic \cite{bhatia_survey_2010}.
Images, on the other hand, are rigid and ordered in their matrix-grid structure, and for a single pixel, its adjacent neighbors can be computed effortlessly. Convolutional neural networks (CNNs) rely heavily on this regularity to classify and segment images. Unlike in 2D, convolutional networks cannot be used to process 3D point cloud data easily, as this type of data lacks explicit spatial ordering. Any algorithm that processes spatial data must operate with strict permutation invariance to guarantee deterministic outputs regardless of the ordering of the input sequence.

In the context of automotive perception, this unstructured data mostly stems from LiDAR sensors, which measure the distance between the vehicle itself and the surrounding space either through the mainstream method of time-of-flight measurement or through a frequency-modulated laser signal \cite{li_lidar_2020}.
Autonomous LiDAR systems serve as a sensory backbone of autonomous driving, delivering a continuous stream of complex geometrical structures, which are susceptible to range-dependent sparsity and jitter through atmospheric processes. 

\subsection{Spatial Benchmark Datasets}

\subsubsection{Real-World Datasets: KITTI and SemanticKITTI}
The KITTI dataset is conceived as one of the first publicly available benchmarks for autonomous driving \cite{geiger_are_2012, geiger_vision_2013}. 
This dataset established a baseline for 3D object detection and tracking, offering raw real-world LiDAR data with corresponding grayscale and color images and GPS/IMU information. The data was captured in the urban environment of a  mid-sized German city as well as rural areas and on highways. A Velodyne HDL-64E rotating LiDAR scanner was mounted on a station wagon, capturing more than one million points per second. More than 200,000 objects were manually annotated with 3D bounding boxes -- like cars, pedestrians, and cyclists -- for ground truth comparison. 
Despite its immense utility, KITTI also introduced architectural challenges.  
Because a rotating LiDAR scanner was used, the resulting point clouds suffer from range-dependent sparsity and directional geometric distortion. As distance increases, the fixed resolution introduces a spatial stretching effect that leaves distant objects represented by only a handful of scattered points.
Furthermore, real-world data acquisition is immensely plagued by long-tail class imbalances, meaning cars and roads occur far more often than motorcyclists or trucks. 

\begin{figure}[ht]
    \centering
    \begin{subfigure}{.5\linewidth}
    \centering
    \includegraphics[width=.9\linewidth]{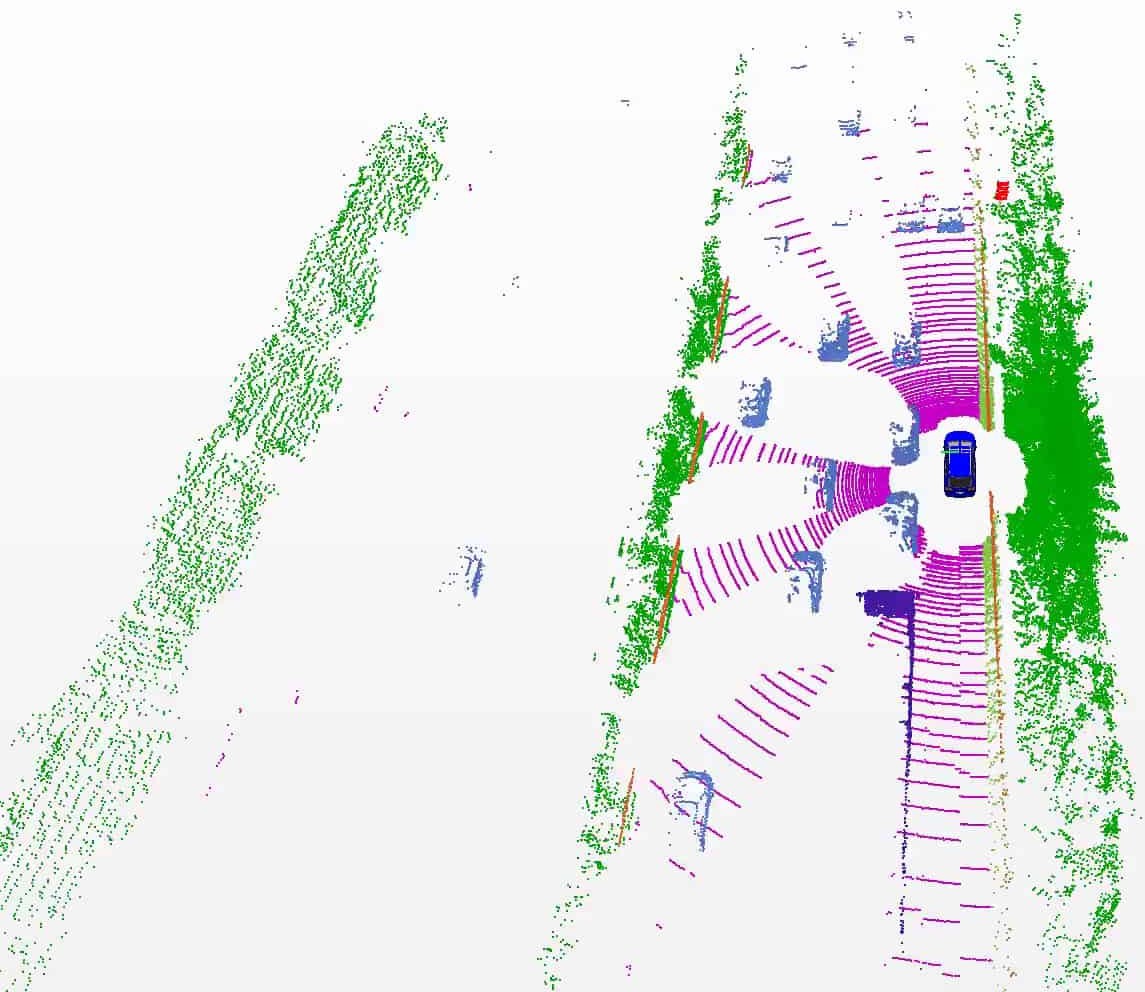} % 
    \caption{ }
    \label{fig:KITTI_ModelNet_sub1_KITTI}
    \end{subfigure}%
    \begin{subfigure}{.5\linewidth}
    \centering
    \includegraphics[width=.9\linewidth]{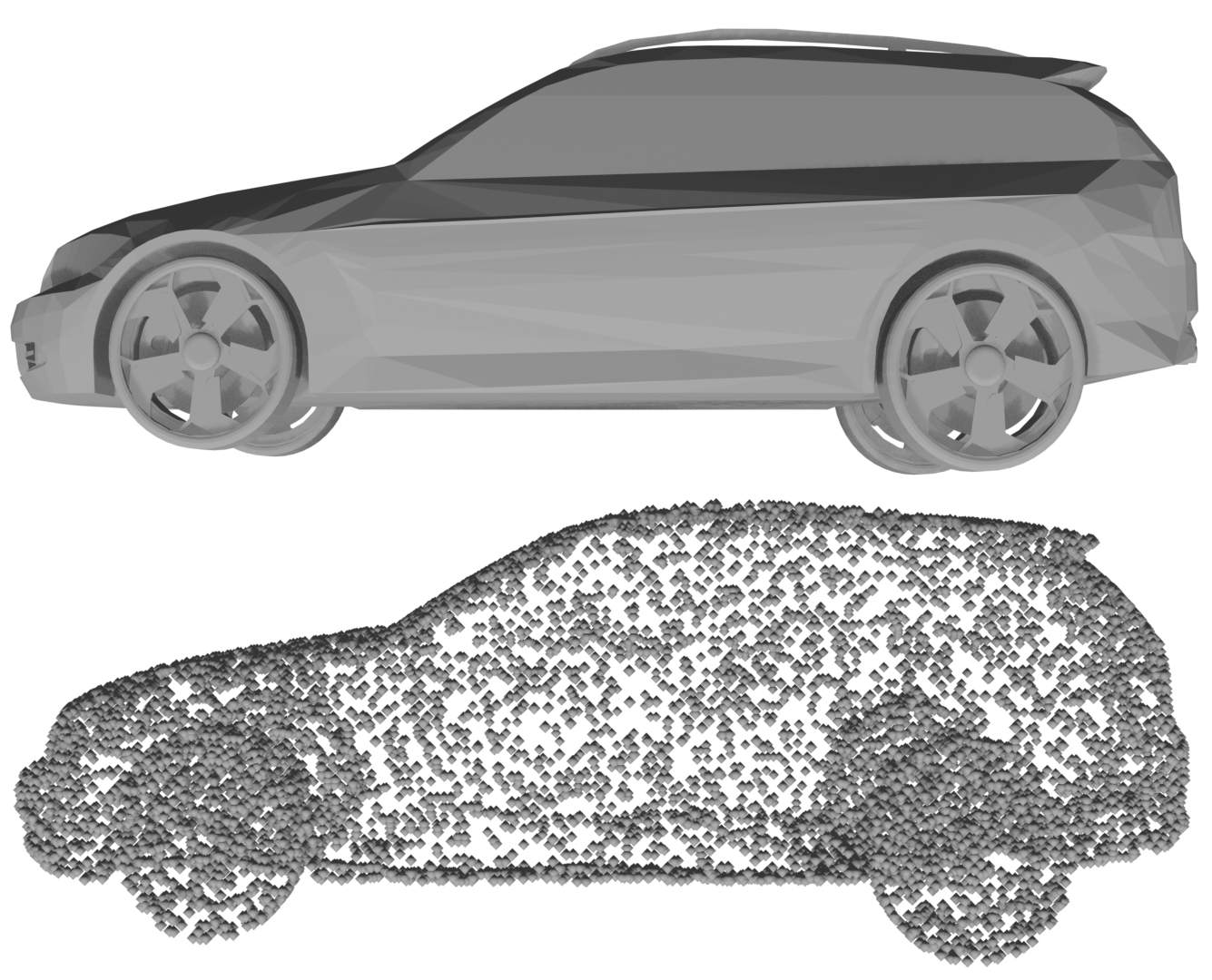} % width=.4\linewidth
    \caption{ }
    \label{fig:KITTI_ModelNet_sub2_ModelNet}
    \end{subfigure}
    %\missingfigure{\textbf{\textcolor{red}{Kitti raus, und ggf. etwas ganz Allgemeines zu Punktwolken}}}
    \caption{Contrast between an instance of \textbf{(a)} SemanticKITTI and \textbf{(b)} ModelNet instance. In \textbf{(b)}, a point cloud is displayed below the actual CAD model. The SemanticKITTI image was taken from \url{https://semantic-kitti.org/images/mobile_hero2.jpg}.}
    \label{fig:semanticKITTI_ModelNet}
\end{figure}

To  extend the utility of the KITTI benchmark even further, Behley et al. introduced a follow-up dataset: SemanticKITTI \cite{behley_semantickitti_2019} (Figure \ref{fig:KITTI_ModelNet_sub1_KITTI}). This dataset draws from the raw data of the KITTI odometry dataset and expands it to a point-by-point annotation, meaning that over 4.5 billion points were annotated into 28 classes. Thus, capturing sequence consistency between the recorded scans. By mapping annotations directly on the unstructured raw dataset, it became an industry standard for training and validation in the field of 3D semantic segmentation, demanding substantial processing power and introducing a severe bottleneck on edge devices. 

\subsubsection{ModelNet: Synthetic Shape Repository}
In contrast to the complex environments captured by real-world automotive LiDAR datasets, the development of 3D deep learning architectures frequently relies on synthetic shape repositories.
The most prominent and widely adopted of these datasets is ModelNet, which was introduced by Wu et al. \cite{wu_3d_2015} (Figure \ref{fig:KITTI_ModelNet_sub2_ModelNet}). Rather than capturing real-world instances of objects, ModelNet consists of clean and geometric flawless 3D CAD models, representing a wide variety of categories. 
Two subsets are open for usage: ModelNet10 and the more challenging ModelNet40 repositories, which are comprised of ten and forty categories, respectively. The ModelNet40 subset is comprised of 12,311 distinct shape instances, ranging from categories like 'airplane' and more organic forms like 'person' and 'flower pot'. In standard literature architectures, instances of these CAD point clouds are preprocessed via uniform sampling and downsampled so that each instance contains exactly 1024 points. 
As this synthetic framework offers models without any range-dependent point drop, sensory jitter, or object occlusion, it allows pipeline architectures to focus on global classification evaluation. 
However, evaluating networks exclusively on ModelNet is not without prominent issues. Recent studies have revealed data irregularities within the repository itself, including duplicated instances, multiple objects that are comprised in one instance, and mislabeling \cite{van_den_herrewegen_point_2023, saeid_enhancing_2025}. Furthermore, because data stem from virtual CAD models and not from sensory input, they do not simulate classical LiDAR behavior. The point clouds completely lack intensity reflectivity values, and as the entire surface is sampled uniformly, they omit occlusions in general. This highlights the gap between synthetic data, which is drawn from CAD shapes, and actual sensory input.  

\subsubsection{Data Augmentation Techniques in Spatial Computing}

To prevent neural networks from overfitting, as not enough data is present in the dataset, data augmentation serves a critical role in the pipeline. Broad surveys by Zhu et al. classify these techniques into basic spatial manipulation and specialized deformations \cite{zhu_advancements_2024}. 

\begin{itemize}
    \item \textbf{Basic Augmentations}: These methods rely on standard affine transformations -- such as random scaling, translation, and rotation -- alongside point dropping and the induction of jitter (Gaussian noise). The networks are forced to become more invariant to minor global or local spatial perturbations. 
    \item \textbf{Specialized Augmentations}: Advanced frameworks implement further strategies, such as PointCutMix, which blends geometric features from different object classes to challenge the networks even further \cite{zhang_pointcutmix_2021}. 
\end{itemize}

The necessity of these techniques is highlighted by frameworks like ModelNet-C, which demonstrate that 3D neural networks are especially sensitive to object orientation \cite{sun_benchmarking_2022}. Augmentation techniques can help to elevate underrepresented classes or challenging instances. 
However, while traditional augmentations can simulate structural variance, they cannot replicate physical sensor phenomena -- such as occlusions or surface reflectivity -- underscoring the need for advanced simulation tools like BLAINDER.

\subsection{Perceptual Neural Network Paradigms for 3D Understanding}
\label{sec:nn_paradigms_3d_understanding}

The landscape of 3D deep learning contains three primary categories, each operating at different levels of spatial and semantic granularity. 

\begin{itemize}
    \item \textbf{3D Point Cloud Classification}: This global task focuses on predicting a single semantic label for an entire point cloud. It serves as the foundational architectural baseline for evaluating feature extraction capabilities. A more detailed explanation is provided in Section~\ref{sec:deep_dive_3d_classification}.
    \item \textbf{3D Object Detection}: Moving beyond global shape recognition, object detection aims to find and categorize specific objects within a point cloud or point cloud sequence. This is typically achieved by drawing 3D bounding boxes and thus predicting an object's spatial measurements, like translation and volumetric dimensions. Modern approaches like VirConv leverage multimodal fusion of LiDAR sensor inputs and camera data \cite{wu_virtual_2023}.  
    \item \textbf{3D Semantic Segmentation}: Operating at the highest level of granularity, semantic segmentation is assigning a specific class label to every individual point within the point cloud itself.  Surveys on large-scale scene understanding \cite{betsas_deep_2025, halperin_point_2025} provide a taxonomy in segmentation strategies: \textbf{projection-based models} (such as SqueezeSeg \cite{wu_squeezeseg_2017, wu_squeezesegv2_2018} and 3D-MiniNet \cite{alonso_3d-mininet_2021}), which map 3D points onto a spherical or cylindrical 2D grid to discretize the unordered set  and accelerate inference, and \textbf{point-based frameworks} (such as RandLa-Net \cite{hu_learning_2022}), which deal with raw unstructured point clouds directly using localized aggregation. Because assigning labels to billions of points is extremely expensive, these tasks introduce severe computational bottlenecks.
\end{itemize}

\subsubsection{Deep Dive into 3D Classification Methodologies}
\label{sec:deep_dive_3d_classification}

To process the unordered and unstructured nature of point sets, classification networks have brought forth several core methodologies, each introducing several trade-offs in their computational efficiency and geometric fidelity. The next section can just deliver a brief overview of current methods. For further study, the following paper by Zhang et al. is recommended \cite{zhang_deep_2023}.

\paragraph{Multi-View and Projection-Based Methods}
Multi-view architectures circumvent the unstructured nature of 3D point cloud data by projecting the raw coordinates onto one or multiple predefined 2D planes, and therefore transforming the problem into the image-based domain. Once projected, the image can be categorized by highly efficient 2D CNNs. This strategy reduces the processing anomalies caused by varying point cloud densities. 

However, multi-view methods require heavy geometric preprocessing and introduce irreversible depth information loss during projection.
To reduce the risk of perspective occlusion, the network needs to rely on an array of camera viewpoints, elevating the computational cost even higher.  

\paragraph{Voxel-Based Methods}
Voxel-based methods solve the challenge of unstructured data by transferring the continuous 3D coordinate space into a regular and rigid, three-dimensional volumetric grid. A volumetric element in this grid is known as a voxel. Once voxelized, a standard 3D CNN can be used to learn neighbor spatial relationships due to the regularized layout. This method was also used  in 3D ShapeNets \cite{wu_3d_2015}.  

All architectures that use voxelization are furthermore bounded by their low grid resolution, which introduces severe quantization artifacts and structural detail loss. If multiple points fall within a single grid element, they are merged into a single voxel, discarding local point counts \cite{tang_searching_2020}. This voxel resolution can limit the detection of small and local features. Furthermore, voxel-based methods suffer from cubic growth in computational and memory complexity relative to the grid resolution. 

\paragraph{Point-Based Methods}
Point-based methods were first pioneered by PointNet \cite{qi_pointnet_2017}. These methods process raw, continuous coordinates directly without prior spatial transformation or preprocessing, thereby fully preserving the underlying 3D geometry. To handle an unordered set of points, PointNet first feeds each point through a shared multi-layer perceptron (MLP) and extracts point-wise features. These features are then aggregated in a subsequent step with the help of a max-pooling function. Max-pooling was chosen as it is a symmetric function (i.e., $f(x,y) = f(y,x)$) and therefore strictly permutation-invariant. 

While PointNet successfully extracts global features, it is limited by its ability to capture local structures. To resolve this, PointNet++ \cite{qi_pointnet_2017-1} introduced hierarchical grouping layers that process a localized neighborhood search via the k-Nearest Neighbors (k-NN) algorithm. However, these local search algorithms are highly resource-intensive. 
Furthermore, point-based methods are density-sensitive. If too many points are clustered in one area of the instance, the classification accuracy quickly diminishes. 

\paragraph{Graph-Based Methods} 
Graph-Based Convolutional Neural Networks (GCNs) treat the point cloud as a graph, where each 3D point serves as a vertex (node), and topological relationships between adjacent points are represented by graph edges. One baseline for this methodology is the Dynamic Graph CNN (DGCNN) \cite{wang_dynamic_2019}. The underlying structure makes the network highly invariant to affine transformation. However, constructing and updating the graph requires dynamically executing k-NN searches, which immensely increases the computational overhead, which is a significant barrier for edge hardware. 

\paragraph{Further emerging trends}
Recent advancements in 3D classification are divided between Transformer-Based and highly optimized networks:

\begin{itemize}
    \item \textbf{Transformer-Based Models}: Frameworks such as Point Transformer \cite{wu_point_2024} and Point2Vec \cite{knaebel_point2vec_2023} use self-attention mechanisms to capture spatial dependencies and complex structures. But their usage of transformers builds up further computational complexity, which diminishes their usage on edge devices.
    \item \textbf{Lightweight and Non-Parametric Models}:  Over the last couple of years, further approaches have been tested, which focus on reducing parameters and improving algorithm efficiency. Frameworks like Point-GN completely omit learnable weights as parameters, instead using static Gaussian positional encoding \cite{mohammadi_point-gn_2024}. Point-SkipNet, on the other hand, heavily uses skip connections, allowing the network to bypass redundant feature layers \cite{saeid_enhancing_2025}. 
\end{itemize}

\subsection{Explainability Methods in 3D Deep Learning}
\label{sec:related_works_explainability}
As neural network architectures grow more and more complex, they fundamentally function as "black boxes" where it is not clear why a certain instance was classified exactly in this way and not in another. In safety-critical domains such as autonomous driving and medical diagnosis, it is key to establish trust in the models. One cannot blindly rely on the model's high validation accuracy but must also understand how the underlying architecture arrives at this geometric interpretation (e.g., identifying trains based on nearby tracks or on the vehicle itself). To open up these black boxes, two primary classes of local explainability methods have emerged: 

\begin{itemize}
    \item \textbf{Gradient-Based Methods}: Frameworks introduced by Zheng et al. \cite{zheng_pointcloud_2019} and by Levi and Gilboa \cite{levi_fast_2024} try to explain the underlying PointNet model by calculating a feature gradient during the backward pass, through systematically shifting points towards the centroid of the model instance to isolate their structural importance. These methods are computationally very efficient as they only require a single backward pass. However, they are strictly model-dependent, meaning they can only be applied to fully differentiable architectures. 
    \item  \textbf{Local Surrogate Based Methods}: Approaches such as LIME-3D are model-agnostic, meaning they treat the underlying network as a "black box" \cite{tan_surrogate_2021}. They group the model into local clusters and generate perturbed variants by removing clusters and thus observing the changes in the network's output classification. Therefore, these methods are highly interpretable for humans but suffer from immense computational cost and "contribution neutralization", where positive and
    negative point activation in one cluster cancel each other out. 
\end{itemize}

Ultimately, these local explainability methods excel at the point-wise individual level, but they are fundamentally limited as they cannot globally explain whole semantic classes. For explaining classes globally, other methods already exist \cite{tan_visualizing_2023}.

\subsection{Methods in Point Cloud Downsampling}
In real-world applications, 3D point clouds can be comprised of millions of points, especially when they are acquired from LiDAR scanners. This introduces an immense data pipeline, which often exceeds edge hardware capabilities. Downsampling serves as a vital geometric preprocessing phase designed to systematically drop points from the initial point set in order to reduce memory consumption and overall execution time while maintaining high accuracy.
Literature primarily divides downsampling methodologies into three distinct algorithmic approaches:

\subsubsection{Farthest Point Sampling}
Farthest Point Sampling (FPS) serves as a foundational structural block in architectures like PointNet++ and SkipNet \cite{qi_pointnet_2017-1,saeid_enhancing_2025}. It is a geometry-driven, iterative selection algorithm. Given an initial point set of $N$ points, the algorithm first selects a starting seed at random. Afterwards, it iteratively populates a subset of $M$ points by calculating the Euclidean distance between all unselected points and the last selected point from the subset, choosing the single point with the highest absolute distance from the current one. 
This distance-maximizing logic guarantees exceptional spatial coverage across the entire surface of the given point cloud. However, because each successive point requires updating a distance measurement to all other points that are not in the target set, FPS suffers from quadratic computational complexity ($O(N\times M)$). When scaling to larger scenes with millions of points, this algorithm can quickly become a bottleneck in execution on edge devices. 

\subsubsection{Random Sampling}
In stark contrast to the heavy geometric computational method of FPS, Random Sampling (RS) uses a statistical downsampling heuristic. The algorithm gives each point in the $N$-sized point set array an equal probability of being chosen, creating a uniform distribution, and then taking $M$ points. Therefore, RS is exceptionally efficient, as it requires zero neighborhood calculations, which makes it highly attractive for real-time edge applications, like RandLA-Net \cite{hu_learning_2022}. 

However, this simplicity introduces major trade-offs because the selection is random and therefore entirely decoupled from the geometry. RS can discard highly important geometric features or only select a localized cluster of points. Furthermore, it is non-deterministic, meaning the output from the same point changes with every execution, which can lead to instabilities.

\begin{figure}[ht]
    \centering
    \includegraphics[width=0.95\linewidth]{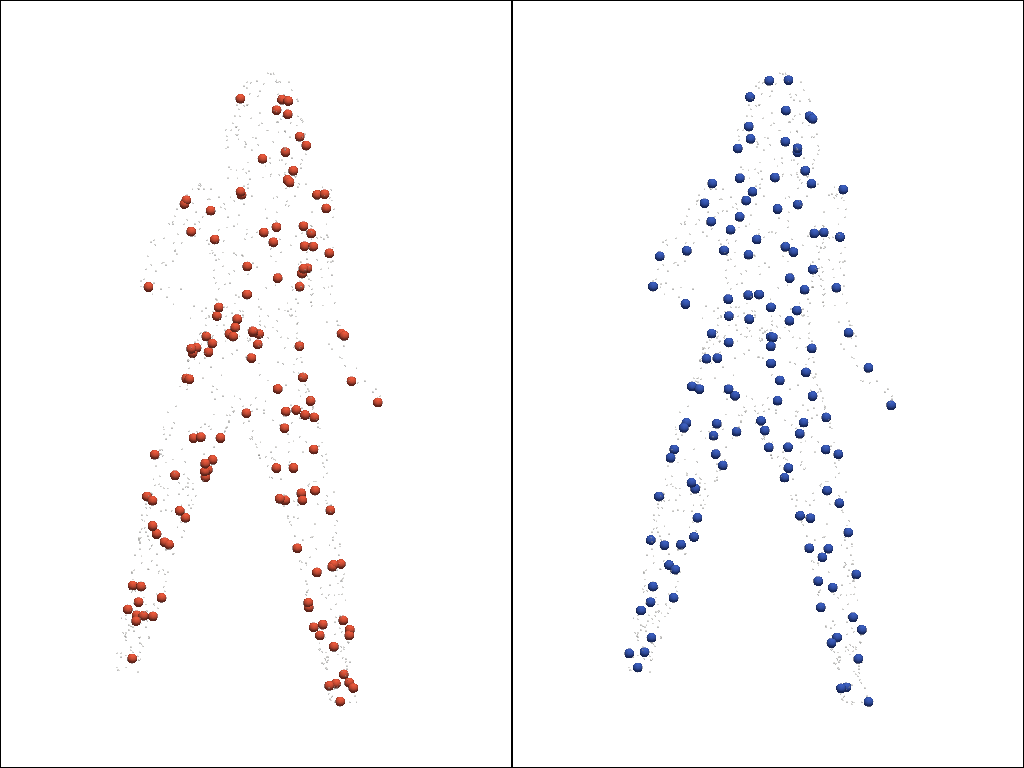}
    \caption{Comparison between a point cloud consisting of 1024 points sampled to 128 points by RS in red on the left and FPS in blue on the right. It can be seen that FPS samples the point set spatially more evenly.}
    \label{fig:rs_fps_comparison}
\end{figure}

An example of RS and FPS can be seen in Figure \ref{fig:rs_fps_comparison}.

\subsubsection{Feature-Based Downsampling: Critical Point Layer}
\label{sec:related_works_cpl}
To merge the deterministic geometric selection of algorithms like FPS and the high execution speed of heuristic methods, Nezhadarya et al. introduced the Critical Point Layer (CPL) \cite{nezhadarya_adaptive_2020}. CPL avoids neighborhood searches and distance measurements entirely by leveraging a shared MLP to project the raw point coordinates into a high-dimensional feature space. The layer then executes a column-wise global max-pooling on the resulting feature matrix, saving the row indices that achieve the highest activation. 
The selected indices are then sorted by the number of channels they dominate via their maximal activation. Therefore, the CPL achieves strict permutation invariance and deterministic inference. 
The primary trade-off of CPL is its explicit task-dependence. If trained alongside a classification network like PointNet, the layer optimizes its internal weights to preserve only features that are necessary to distinguish between object classes and discards all other points that achieve a low activation in its MLP. 
\section{Methodology}
\label{sec:methodology}

To bring 3D deep learning together with edge devices, one needs to develop a specialized hardware-constrained workflow. State-of-the-art spatial perception models rely on heavily parameterized networks and high-end GPUs, which are often unavailable for local robotics and other off-the-grid approaches.
This work focuses on the affordable and hardware-constrained edge platform: the Raspberry Pi 5. Through the use of the aforementioned hardware, we want to put existing methods to the test. 
Prior benchmarks that were conducted by Wisultschew et al. used an earlier generation of the Raspberry Pi family -- namely the Raspberry Pi 4 \cite{wisultschew_characterizing_2022}. Their evaluation demonstrated that the Raspberry Pi 4 could already execute point cloud classification successfully after optimization -- achieving an inference time of  $236.65$ ms when deploying PointNet directly.
Leveraging the capabilities of the newer ARM Cortex-A76 processor of the Raspberry Pi 5, we want to assess the latency reduction achievable under edge constraints using the latter described methodology.

\begin{figure}[tb]
    \centering
    \includegraphics[width=1.\linewidth]{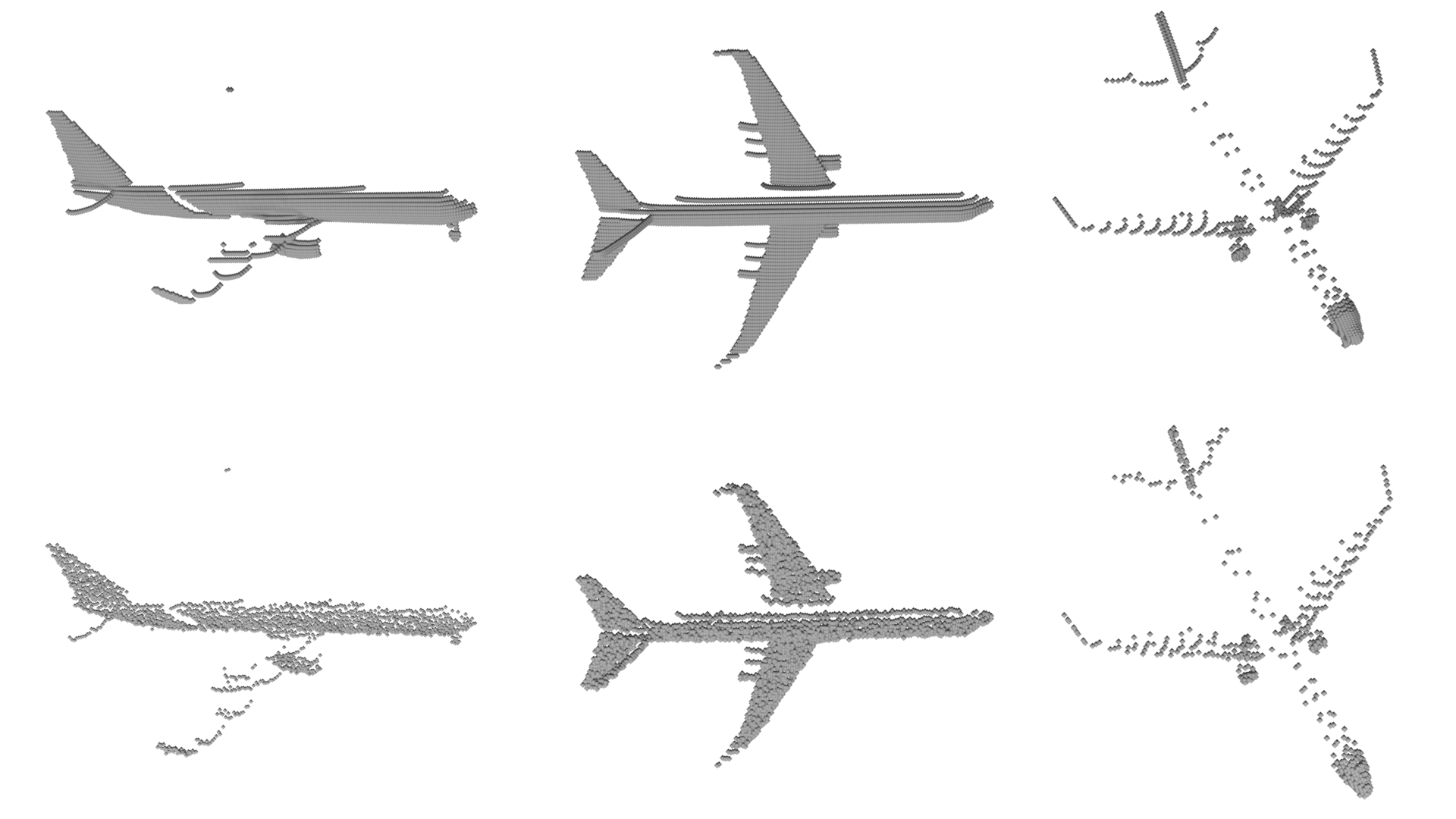}
    \caption{Synthetic rotational LiDAR scans of instance \texttt{airplane\_0655} recorded by BLAINDER. The records in the top row do not inherit noise, whereas those in the lower rows do. From left to right, the columns depict a right-side lateral view, a top-down view, and a frontal view of the aircraft.}
    \label{fig:syth_lidar_rot}
\end{figure}

\begin{figure}[tb]
    \centering % 
    \includegraphics[width=0.95\linewidth]{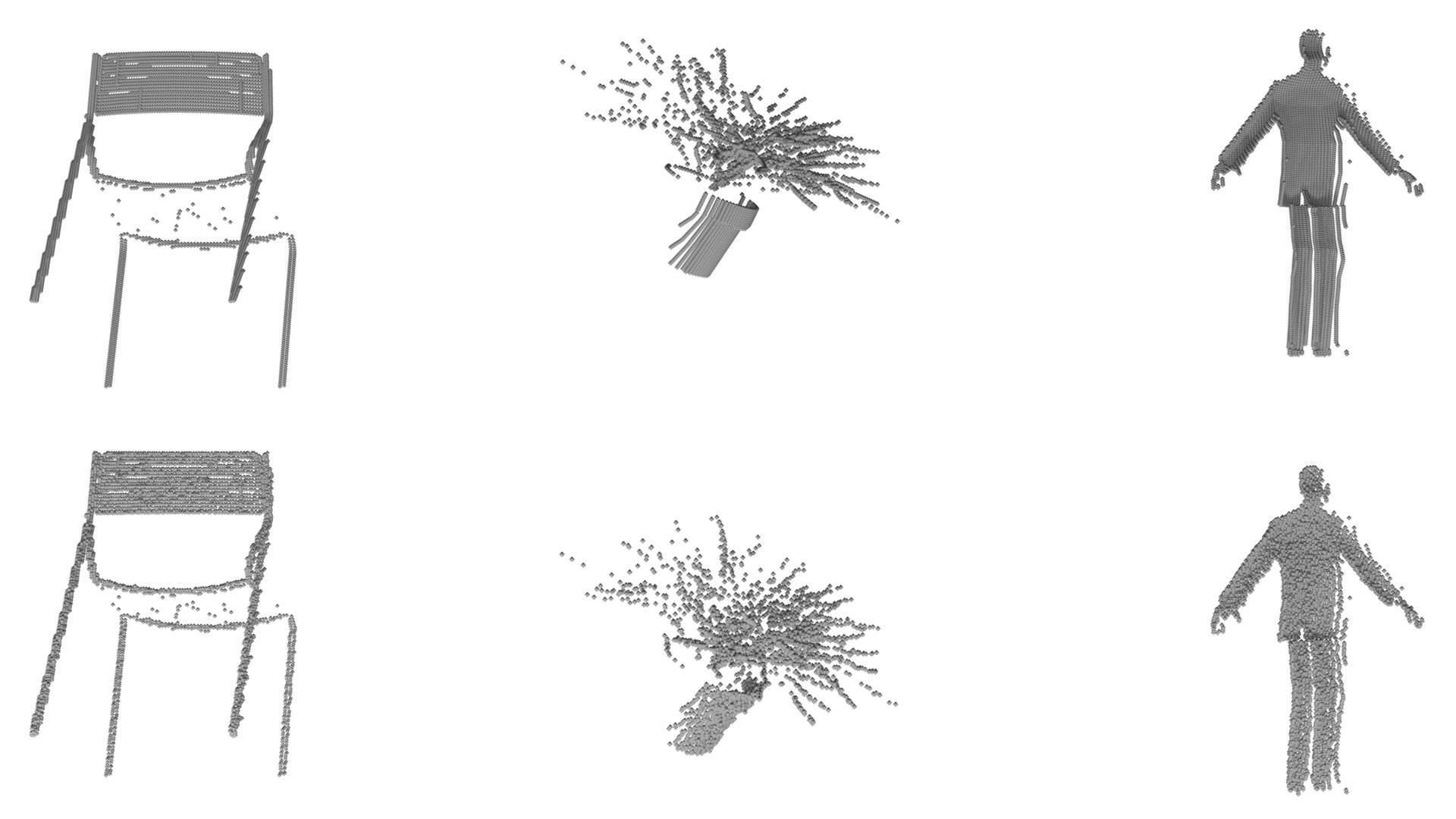}
    \caption{Synthetic LiDAR scans by a generic static scanner. The upper row displays \texttt{chair\_0006}, \texttt{flower\_pot\_0029}, and \texttt{person\_0027} without any induced noise; the lower row shows the instances with induced noise.}
    \label{fig:synth_lidar_generic}
\end{figure}

\subsection{Bridging the Sensor Reality Gap}

While the aforementioned existing implementations demonstrated the viability of deploying PointNet on edge hardware, their evaluation relied exclusively on uniformly sampled instances from the ModelNet dataset. These instances are completely free of environmental noise. To close this existing gap, this workflow leverages BLAINDER, an open-source add-on for Blender designed for depth-sensing simulation, such as Sonar and LiDAR. By using physically based raycasting methods, instead of simple primitive coordinates, we construct a robust dataset for network training and validation. It inherits geometrical noise and imperfections. 

Our primary goal was to automatically generate synthetic LiDAR equivalents of the ModelNet40 repository. A similar initiative was recently done by the ModelNet-C benchmark. It analyzed how different types of networks tolerate noisy data as well as corrupted representations \cite{sun_benchmarking_2022}. While ModelNet-C also uses synthetic LiDAR scans, it does not inherit any imperfections or noise. Furthermore, realistic representations of real-world scanners were not used. 
To establish a more realistic portrait of the physical world, we captured data from multiple orientations around each object -- as done in ModelNet-C as well. 
To ensure comparative continuity, we utilize PointNet as the core classification backbone throughout this work. It serves as a direct baseline in Wisultschew et al., and the ModelNet-C corruption benchmarks. PointNet remains the foundational, widely accepted standard for evaluating 3D deep learning pipelines.

\subsubsection{Generational Dataset Presets}
To thoroughly evaluate the classification boundaries of PointNet, we derived two distinct datasets that are publicly available to reproduce and hopefully improve our results: \footnote{The dataset \cite{Meyer2026SyntheticLiDAR} can be found here: \url{https://doi.org/10.5281/zenodo.21835460}}

\begin{itemize}
    \item \textbf{Rotational LiDAR Dataset}: This first part of the dataset uses BLAINDER's built-in emulator for the Velodyne UltraPuck scanner, adopting its native parameters to mimic an industry-standard automotive sensor. Furthermore, this configuration has two training sets: a clean, noise-free set and a corrupted set with pointwise injected noise, specifically Gaussian noise with zero mean and a standard deviation ($\sigma$) of 0.01. This is to evaluate the exact effect of noise on the classification accuracy. Every instance was scanned 4 times from different directions. Through this, the neural network should learn not just one view of the object but gain a more holistic understanding (see Figure \ref{fig:syth_lidar_rot}). 
    \item \textbf{Generic LiDAR Dataset}: Serving as a control group, this dataset uses the generic preset presented by BLAINDER, fixed on a single perspective. Here, a "clean" and a "noisy" subset exist as well, as in the rotational dataset (see Figure \ref{fig:synth_lidar_generic}). 
\end{itemize}

As the data-generation logic is built upon an adaptive script, it is totally possible to construct tailored solutions. Adaptations in the pipeline can be made to incorporate weather-dependent noise, such as rain, or to drive up mean and standard deviation, or swap sensor profiles completely -- depending on the target application.   

\subsubsection{Dataset Challenges}

The resulting multi-view datasets introduce structural challenges that test how well PointNet generalizes beyond the standard point cloud representations. One hurdle to be overcome by the network is semantic ambiguity, which is caused by different viewpoints. Because the camera captures an instance from multiple sides, some instances from distinct classes can look practically identical. For example, a folded laptop scanned directly from above has a planar surface profile, just like a dining table.

Furthermore, data anomalies arose during the generation and capturing of the data. As objects inherit different sizes and profiles, and therefore the placement in the virtual camera could change from one instance to the next, this meant that point cloud sizes varied in size. Occasionally, objects were just partially captured, or no point at all was generated. These instances can be excluded during a data preprocessing stage  in a threshold filter. If an instance fails to meet the minimum required point count, it will not be used in the training or testing stage. On the other hand, scans with thousands or tens of thousands of points were uniformly downsampled to maintain consistency in the number of overall points.

While this BLAINDER pipeline captures the geometry quite close to a real LiDAR scanner, it does not capture intensity values.  In real-world datasets, a LiDAR sensor captures intensity values, which depend on the material of the object that is hit by the laser. 
In further iterations of the code, this can be addressed, as BLAINDER is capable of capturing these intensity values. Then, materials would need to be hand-picked for every object, as intensity values need to  be matched to the real-world ones. 

\subsection{Feature-driven Data Thinning: Critical Point Layer}
\subsubsection{PointNet's Critical Point Set}
To extract features in the PointNet architecture, the points are first processed independently in a shared MLP to extract further features. These features are then processed through a global max-pooling operation. A key insight of the authors was that a model's final classification capability solely depends on a small subset of the input data -- the Critical Points. These critical points have the highest activation in the max-pooling operation. Furthermore, they represent global geometric boundaries and landmarks of the shape. 

While dropping up to $50\%$ of the input points, the authors noted that the classification performance only dropped by $3.8\%$ when using random sampling and thus still achieved an adequate classification performance. This proved that PointNet is highly robust to missing data, as long as critical points are not discarded. Because one point can dominate many columns in the max-pooling operation, the actual number of unique critical points that are driving the classification is quite small.

\subsubsection{The Downsampling Dilemma}
This leads to the question if it is possible to extract these critical points early on in the pipeline by utilizing a single downsampling technique. This would reduce computational resources -- especially on edge devices. 
While explainability methods like LIME-3D (see Section~\ref{sec:related_works_explainability}) give an intuitive understanding of which points are more important than others for the classification itself, they cannot be used for downsampling methods. They present a fundamental chicken-and-egg problem: a surrogate explainer requires a complete forward pass and a known classification result to determine the importance of point clusters. Consequently, they cannot be used as a downsampler, as the network does not know which class the instance belongs to beforehand. 
It would certainly be possible to construct a hand-crafted heuristic downsampler that scans for geometric extrema (e.g., maximum and minimum coordinate values). However, these heuristics would depend on the dataset itself. Engineers must manually analyze each class individually to identify which point can be dropped. This method could not be generalized.  

\subsubsection{CPL as Geometric High Pass Filter}

To achieve a general, deterministic, and computationally efficient downsampling technique, this method uses the Critical Point Layer (CPL). CPL functions as a geometric high-pass filter, keeping structural landmarks while filtering out redundant points and point regions (see Section~\ref{sec:related_works_cpl}). 
The layer mirrors PointNet's max-pooling mechanism. Coordinates are processed by a shared lightweight MLP to map them to a higher feature space and execute a global max-pooling step. The argmax indices with the highest activation are kept, while others are dropped. Since one point can dominate multiple columns in the feature matrix, duplicates are eliminated, creating a unique set of indices. At last, these indices are sorted by their feature activation, ensuring that the sampler remains invariant to permutation.   

\begin{figure}
    \centering
    \includegraphics[width=0.65\linewidth]{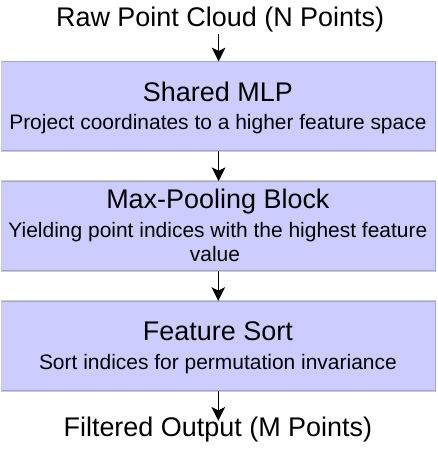}
    \caption{CPL pipeline for point cloud downsampling.}
    \label{fig:cpl_explanation}
\end{figure}

While Nezhadarya et al. designed an entire network architecture around the CPL block integrated with components from other frameworks, our methodology treats the layer as an isolated, standalone filter (see Figure \ref{fig:cpl_explanation}). 
To realize this, we had to implement a three-step plan: 

\begin{enumerate}
    \item \textbf{Fusion}: We fuse a trainable MLP and the CPL block directly onto the front-end of our PointNet classification model, creating a unified network. 
    \item \textbf{Training}: The fused network is trained together on the ModelNet40 dataset, to be easily comparable to other existing downsampling methods. As the entire pipeline is differentiable, the classification loss backpropagates through PointNet and forces the CPL front-end MLP weights to optimize for classification. The MLP learns which points are class-defining and belong to structural landmarks and which do not. 
    \item \textbf{Extraction}: Once the whole network is trained, we isolate the CPL front-end from the initial classification head, freeze the learnable parameters, and export it as a standalone module.
\end{enumerate}

The end-to-end trained front-end can then be deployed on the Raspberry Pi to compress raw point clouds to a denser representation with only a fraction of the initial point cloud size.  
\section{Evaluation and Results}
\label{sec:results}

\subsection{Setup and Baseline Evaluation}
To quantitatively evaluate the performance of our workflow, we used an available PyTorch implementation of PointNet.\footnote{\url{https://github.com/yanx27/Pointnet_Pointnet2_pytorch}} The network was optimized utilizing the standard hyperparameters from the baseline repository: training was conducted over 200 epochs with a batch size of 24, using the Adam optimizer with a learning rate of 0.001.

All models were trained using an 80/20 train-test split of the ModelNet40 dataset. In theory, the Raspberry Pi 5 would be capable of training such a small network; the limited performance led us to train all models on an external workstation machine. Inferencing was executed on the Raspberry Pi 5 CPU. Furthermore, the implementation supports a multi-vote system to augment the classification results by running the network multiple times sequentially and accumulating the results. This was omitted as it would skew the latency benchmarks. 

The baseline model was trained and evaluated on 1024 points per instance, as this point amount was also used in PointNet. Our trained model achieved an instance accuracy of $91.09\%$ and a mean class accuracy of $86.28\%$. An analysis of the confusion matrix (see Figure~\ref{fig:confusion_matrix_pointnet_modelnet}) revealed classification anomalies that were concentrated between semantically  similar groups, such as 'plant' and 'flower pot'. It would certainly be possible to switch to an edited dataset version, such as ModelNet-R, to yield a higher classification result, but the standard ModelNet40 repository was chosen to ensure a clean comparison against other methods.  

Qi et al.~\cite{qi_pointnet_2017} demonstrated that a model that was trained with a point density of 1024 points can retain good classification accuracy even when up to half of the points are dropped during the test phase. However, it was previously not tested whether training on fewer points causes classification performance to diminish or if it remains adequate for classification tasks. 
Inference on ultra-sparse subsets can be a possible path for resource-constrained edge hardware. To implement this, the dataset was systematically downsampled via uniform random sampling and trained with fewer points.

\begin{table}[htbp]
\caption{Demonstration of the robustness of PointNet to variants of point density in training. The input points indicate the number of points that were used for training.}
\begin{tabular}{l|l|l}
\hline
\textbf{Input Points ($N$)} & \textbf{Instance Acc.} & \textbf{Class Acc.} \\ \hline
1024                        & 91.09 \%                          & 86.28\%                                 \\
512                         & 90.63 \%                        & 87.20\%                                 \\
256                         & 90.19 \%                      & 86.66\%                                 \\
128                         & 89.86 \%                           & 86.65\%                                 \\
64                          & 89.01 \%                           & 84.44\%                                 \\
32                          & 87.41 \%                           & 82.60\%                                 \\
16                          & 83.01 \%                           & 76.72\%                                
\end{tabular}
\label{tab:pointnet_training_fewer_points}
\end{table}

The quantitative results show PointNet's great structural robustness (see Table \ref{tab:pointnet_training_fewer_points}). Even when thinned to 16 points, the network maintains an instance accuracy of 83.01\%, while only using a fraction of the original density. 
This resilience can be used for tailored edge AI deployment: a network can specifically be optimized to achieve near real-time execution while maintaining high accuracy. 

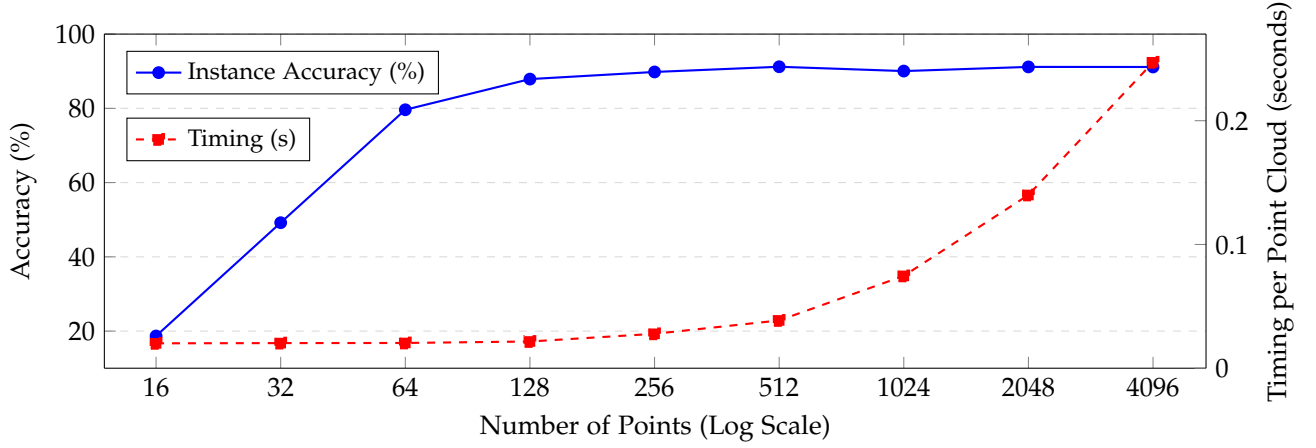
\begin{figure*}[htbp]
\centering
\begin{tikzpicture}
\begin{axis}[
    width=0.95\textwidth, % <--- Stretches the plot horizontally
    height=6cm,           % <--- Keeps a clean, wide aspect ratio
    xlabel={Number of Points (Log Scale)},
    ylabel={Accuracy (\%)},
    xmode=log,
    log basis x={2},
    xtick={16,32,64,128,256,512,1024,2048,4096},
    xticklabels={16,32,64,128,256,512,1024,2048,4096},
    xmin=12, xmax=5500,
    ymin=10, ymax=100,
    axis y line*=left,
    ymajorgrids=true,
    grid style={dashed, gray!30},
    legend style={at={(0.02,0.95)}, anchor=north west, font=\small}
]
\addplot[color=blue, mark=*, thick] coordinates {
    (16, 18.68)
    (32, 49.20)
    (64, 79.62)
    (128, 87.88)
    (256, 89.80)
    (512, 91.20)
    (1024, 90.04)
    (2048, 91.16)
    (4096, 91.14)
};
\addlegendentry{Instance Accuracy (\%)}
\end{axis}

\begin{axis}[
    width=0.95\textwidth, % <--- MUST match the first axis width exactly
    height=6cm,           % <--- MUST match the first axis height exactly
    ylabel={Timing per Point Cloud (seconds)},
    xmode=log,
    log basis x={2},
    xmin=12, xmax=5500,
    ymin=0, ymax=0.27,
    axis y line*=right,
    axis x line=none,
    legend style={at={(0.02,0.75)}, anchor=north west, font=\small}
]
\addplot[color=red, mark=square*, dashed, thick] coordinates {
    (16, 0.0201)
    (32, 0.0203)
    (64, 0.0204)
    (128, 0.0216)
    (256, 0.0278)
    (512, 0.0386)
    (1024, 0.0746)
    (2048, 0.1400)
    (4096, 0.2472)
};
\addlegendentry{Timing (s)}
\end{axis}
\end{tikzpicture}
\caption{PointNet performance profile across different downsampling granularities. The plot illustrates a linear computational complexity relative to point density; note that the curve exhibits a visually exponential trajectory solely due to the logarithmic scaling of the horizontal axis.}
\label{fig:pointnet_downsampling_inference}
\end{figure*}

\subsubsection{Synthetic LiDAR Data Evaluation}
To systematically evaluate the generalization boundaries of our pipeline, a cross-evaluation was conducted across five distinct dataset configurations: the original clean ModelNet repository, the simulated static (clean and noisy) streams, and the simulated rotational (clean and noisy) variants. To ensure a comparable evaluation, independent PointNet models were optimized for each dataset utilizing the baseline hyperparameters.
The trained networks were evaluated on all datasets. The resulting cross-matrix captures both instance accuracy (I) and class accuracy (C) (see Table \ref{tab:network_evaluation}). 

\begin{table*}[htbp]
\centering
\caption{Evaluation of trained network architectures across different test datasets (Accuracy \%). Bold values highlight in-domain baseline evaluations (matching train and test distributions).}
\label{tab:network_evaluation}
\small
\begin{tabular}{l *{10}{r}}
\toprule
\textbf{Test Dataset} & \multicolumn{10}{c}{\textbf{Trained Network}} \\
\cmidrule(lr){2-11}
 & \multicolumn{2}{c}{Original} & \multicolumn{2}{c}{Static Clean} & \multicolumn{2}{c}{Static Noisy} & \multicolumn{2}{c}{Rotation Clean} & \multicolumn{2}{c}{Rotation Noisy} \\
\cmidrule(lr){2-3} \cmidrule(lr){4-5} \cmidrule(lr){6-7} \cmidrule(lr){8-9} \cmidrule(lr){10-11}
 & \multicolumn{1}{c}{I} & \multicolumn{1}{c}{C} & \multicolumn{1}{c}{I} & \multicolumn{1}{c}{C} & \multicolumn{1}{c}{I} & \multicolumn{1}{c}{C} & \multicolumn{1}{c}{I} & \multicolumn{1}{c}{C} & \multicolumn{1}{c}{I} & \multicolumn{1}{c}{C} \\
\midrule
Original       & \textbf{91.09} & \textbf{86.28} & 16.63 & 13.36 & 15.78 & 13.34 & 11.81 & 10.98 & 10.84 & 11.57 \\
Static Clean   &  4.78 & 10.64 & \textbf{71.49} & \textbf{66.59} & 58.77 & 59.06 & 15.66 & 18.23 & 14.82 & 16.57 \\
Static Noisy   &  5.04 & 11.17 & 50.56 & 58.60 & \textbf{73.86} & \textbf{68.40} & 12.19 & 15.82 & 11.50 & 15.07 \\
Rotation Clean &  2.28 &  5.28 &  8.75 &  8.04 &  7.12 &  7.11 & \textbf{66.95} & \textbf{60.09} & 57.35 & 51.17 \\
Rotation Noisy &  4.81 &  6.48 &  8.78 &  7.67 &  7.13 &  7.47 & 57.35 & 51.12 & \textbf{59.10} & \textbf{52.83} \\
\bottomrule
\multicolumn{11}{l}{\footnotesize \textit{Note:} \textbf{I} stands for Instance-level accuracy, and \textbf{C} stands for Class-level accuracy.} \\
\end{tabular}
\end{table*}

This evaluation shows that accuracy collapses when translating between different "sensor" types. Most notably, when the network is trained on the clean baseline ModelNet partition but tested on the derived synthetic LiDAR datasets, classification accuracy plummets into a random distribution profile - falling as low as $2.28\%$ on rotational data. The network was trained on whole surfaces of CAD models, but fails to recognize realistic LiDAR data, where range-dependent density drops and jitter exist. 

Conversely, the data reveals domain abstraction driven by the inclusion of sensor-level noise. Models exposed to point-wise corruption during the training exhibit the capacity to abstract to classify clean datasets. For example,  the network trained on the \textit{Static Noisy} subset, retains an instance accuracy of $58.77\%$
when tested on the non-noise induced equivalent. In contrast, the \textit{Static Clean} architecture undergoes a 21-percent-point performance drop and falls to $50.56\%$ accuracy when testing on the noisy profile. This asymmetry indicates that the injection of noise during the training forces the network to focus even more on structural shapes.

Finally, the cross-matrix highlights a critical structural disconnect between the static and rotational scanning profiles, driven by both sensor mechanics and viewpoint density. Networks optimized on a single static perspective undergo a  classification performance collapse -- rendering them entirely unusable -- when evaluated against the multi-view synthetic rotational Velodyne UltraPuck data ($7\%$ to $8\%$ accuracy). Conversely, the rotational model fails equally when restricted to single-view static inputs ($11\%$ to $15\%$ accuracy). 
This behavior demonstrates that the PointNet architecture does not only bind itself to object geometry. Instead, it relies heavily on the specific spatial data distribution and viewpoint coverage present during training, proving sensitivity to the number of viewing angles.

\subsubsection{Hardware Profiling: Latency Analysis through Downsampling}
To characterize the performance of the Raspberry Pi, a test-time analysis was conducted. The standard network parameters that were trained on 1024 points were used. For every point density count, 5000 point clouds were selected randomly from the test set to measure real-time inference latency (see Figure \ref{fig:pointnet_downsampling_inference}).

From the hardware profiling, it can be concluded that a linear complexity exists between the point density and the processing latency. Because PointNet's shared MLP evaluates individual coordinates independently, reducing the point count directly scales the number of operations. Dropping half of the points also reduces the inference timing by about 50\%. 
However, in the figure, it can be seen as well that under 128 points the timings do not decrease as rapidly anymore -- hitting a baseline of approximately 0.02 seconds. This indicates that at these low densities, the CPU execution time is no longer bound by the number of points in the point cloud but by the rest of the network, as well as structural overhead. 
This latency floor shows an optimization opportunity for edge development. Since processing 128 points takes roughly the same time as processing 16 ($\sim 21$ms), but yields a high performance increase in instance accuracy ($87.88\%$ vs $18.68\%$), it can be concluded that 128 points are a good boundary for test-time heuristics. Evaluating the model every 21ms means that the system achieves about 47 FPS. With these performance figures, a multi-vote system could be used to further boost accuracy figures, without breaking real-time latency bounds. 

% edge performance depends on software environment: testing was done on Python 3.7.12 - migrating to Python 3.10.12, pipeline achieved latency improvement 

\subsection{Latency Evaluation of FPS and RS}
To gain a predefined operation efficiency of the data-thinning methods, a latency profiling between FPS and RS was computed on the Raspberry PI. Several target point sizes were used -- ranging from 16 to 512 points -- to gain an understanding of what the CPL method needs to achieve in order to be competitive. 
For evaluation, the \texttt{airplane\_0655} test instance from the ModelNet40 dataset was selected as a representative and moderately complex sample. 
As FPS and RS operate strictly on spatial coordinates, their execution time is independent of the semantic shape of the object; the runtime is solely driven by the input  and target point densities. To ensure statistical consistency, the profiling was executed 5000 times.  

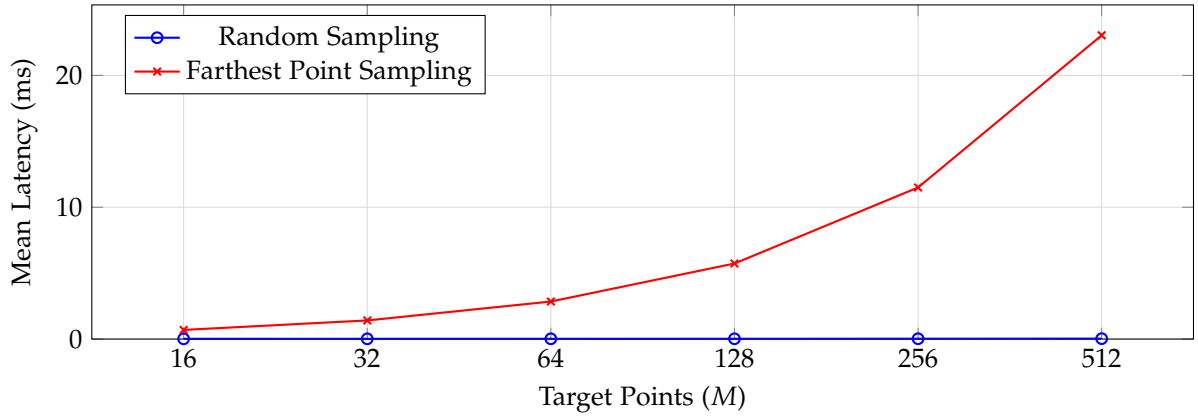
\begin{figure*}[htbp]
    \centering
    \begin{tikzpicture}
        \begin{axis} [
            width=0.95\textwidth,
            height=6cm,
            xlabel={Target Points ($M$)},
            ylabel={Mean Latency (ms)}, % <--- Changed unit to milliseconds
            xmode=log,
            log basis x={2},
            xtick={16,32,64,128,256,512},
            xticklabels={16,32,64,128,256,512},
            ymin=0,                     % <--- Starts axis at 0 for clarity
            scaled y ticks=false,       % <--- Forces pgfplots to never use 10^-2 notation
            grid=both,
            grid style={line width=.1pt, color=gray!10},
            major grid style={line width=.2pt, color=gray!30},
            legend pos=north west,
        ]
        
        % Random Sampling Curve (Converted to ms by multiplying by 1000)
        \addplot[color=blue, mark=o, thick] coordinates {
            (16, 0.016657)
            (32, 0.017351)
            (64, 0.018832)
            (128, 0.020811)
            (256, 0.024580)
            (512, 0.031077)
        };
        \addlegendentry{Random Sampling}
        
        % FPS Sampling Curve (Converted to ms by multiplying by 1000)
        \addplot[color=red, mark=x, thick] coordinates {
            (16, 0.693)
            (32, 1.412)
            (64, 2.847)
            (128, 5.731)
            (256, 11.498)
            (512, 23.042)
        };
        \addlegendentry{Farthest Point Sampling}
        
        \end{axis}
    \end{tikzpicture}
    \caption{Inference preprocessing runtime scaling on the Raspberry Pi 5 architecture. The actual timings for RS are between $0.016657$ms for 16 points and $0.031077$ms for 512 points, whereas FPS is between $0.693$ms $23.042$ms, respectively.}
    \label{fig:latency_comparison_fps_rs}
\end{figure*}

The profiling reveals a stark contrast between heuristic and geometric sampling (see Figure \ref{fig:latency_comparison_fps_rs}). Random sampling execution times are almost instantaneous and constant, ranging from 16.6 µs at 16 points to 31.1 µs at 512 points. On the other hand, FPS displays a greater latency rise due to the underlying quadratic nature of the algorithm. At 16 points, FPS requires 0.693 ms of processing time. When scaling to 512 points, the method requires 23.04ms.

This metric shows a core problem in edge computing: geometric algorithms for the preprocessing of point clouds, such as FPS, can take more time than running a neural network afterwards, and thus create a bottleneck. The goal of CPL is to circumvent this bottleneck. 

\subsection{CPL Training}
The CPL frontend was built up by a pointwise shared MLP, expanding from 3 to 64 over 128 to 1024 channels. Afterwards, a column-wise max-pooling block was used to extract indices with the highest scores. The entire network chain was trained with the Adam optimizer and a learning rate of 0.001 over 200 epochs. 
As the gradient flows strictly through the selected indices during backpropagation, the frontend operates as a discrete bottleneck rather than a continuous function. 

It has to be noted that a single point of a point cloud can inhabit many distinct  features if it is a prominent landmark. For example, a wingtip edge or nose tip can achieve peak activations across multiple feature channels at the same time. Therefore, the unique rows selected by the argmax operator are significantly lower than the channel depth. It is possible that a point cloud with 1024 can be compressed down to as few as 40 points. 

To maintain desired output point sizes, our proof-of-concept implementation duplicates the resulting points. If, on the other hand, fewer points are desired, the number of points is truncated. In later implementations, weighted sampling could be used instead of naive row repetition -- effectively weighting how "valuable" a point is in the network -- and repeat coordinates proportionally. This idea was already proposed by the authors of the publication our method is based on. It would also be possible to execute secondary pooling passes until the desired target density is reached. 

Profiling the isolated CPL frontend over 5,000 test iterations on the same instance as in Figure \ref{fig:rs_fps_comparison} reveals an average execution latency of $1.96$ms for a 128-point target. Compared to the $5.73$ms runtime required by FPS for the identical subset size, our implementation is  nearly three times faster.

\subsection{Geometric and Perceptual Comparison}
\begin{figure*}[htbp]
    \centering
    \includegraphics[width=0.95\linewidth]{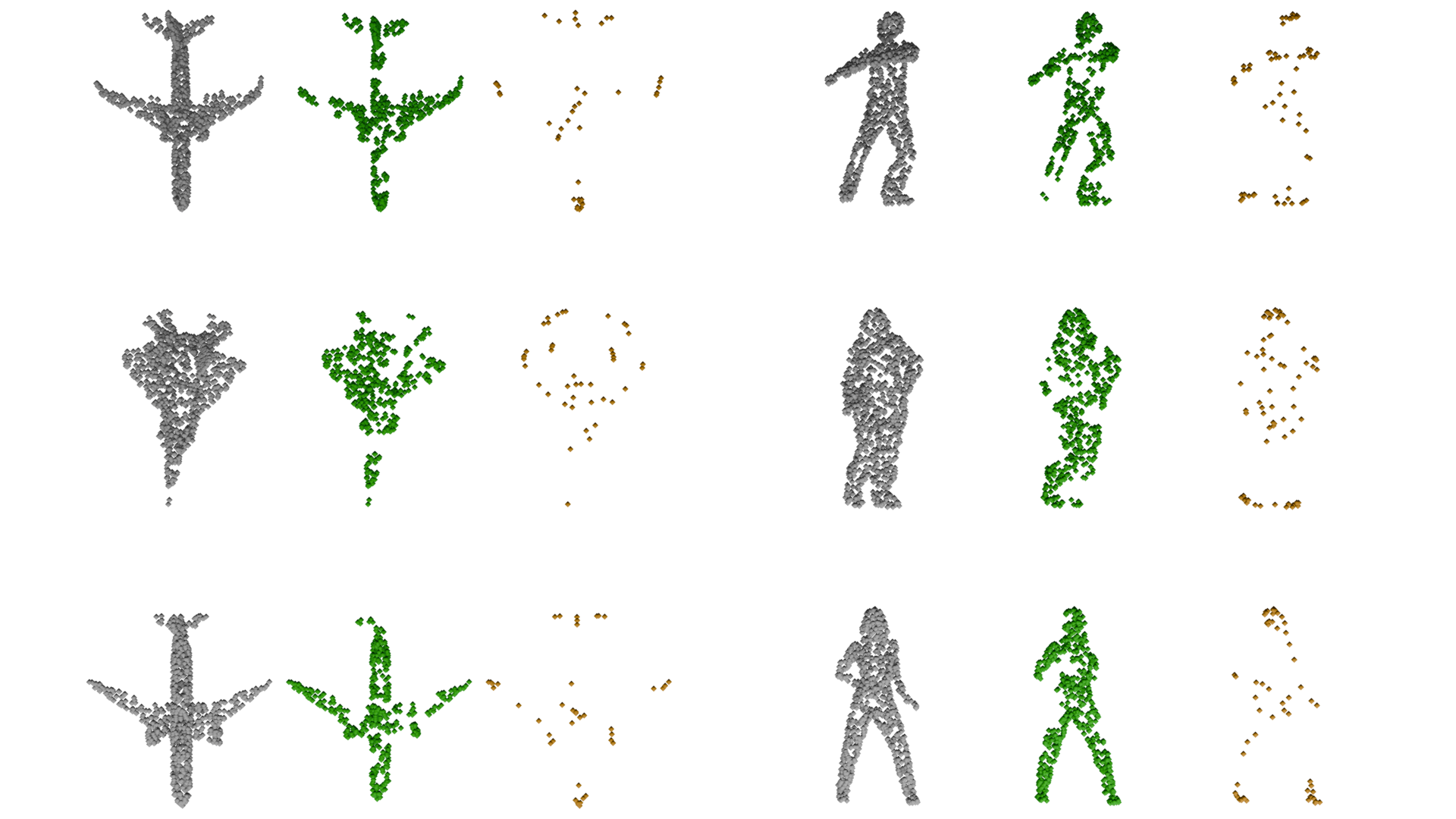}
    \caption{Different airplane and person instances are displayed. The instances in gray are sampled down to 1024 points and taken as input for LIME-3D and CPL. Next to it in green are all positive attributed points, identified via LIME-3D. On the right are the compressed point clouds via CPL in orange.}
    \label{fig:cpl_lime_comparison}
\end{figure*}

To analyze whether the feature-driven downsampler aligns with human perception, we conducted a comparison between several instances from the airplane and person classes of the ModelNet40 dataset. The instances were first randomly sampled to have a point count of 1024, then LIME-3D was applied to see which points positively contributed to the classification. At last, CPL was applied to the downsampled point set. A visual comparison can be found in Figure \ref{fig:cpl_lime_comparison}. It can be noted that CPL learned to isolate extrema and sharp boundaries. For airplane instances, the layer consistently selects points distributed across both wing edges. 

To also quantitatively evaluate the point clouds, a similarity measurement on the basis of the Chamfer Distances (CD) was conducted. Here, the CPL points were compared against the results of LIME-3D as well as a random point cloud with the same number of points as the CPL ones. As the results of RS can vary, we computed the CD 1000 times for RS.   

Across all evaluated classes, RS consistently achieved a lower CD to LIME-3D, maintaining metrics nearly twice as low as CPL (see Table \ref{tab:chamfer_dist_rs_cpl}). This indicates that CPL introduces a higher geometric dissimilarity to the local explainer. 
This highlights the trade-off between classification utility and human interpretability, while exposing boundaries of both frameworks. 

\begin{table}%[htbp]
    \centering
    \caption{Chamfer Distance (CD) Dissimilarity Analysis Relative to LIME-3D Explanations between Random Sampling (RS) -- Evaluated over 1,000 iterations -- and the hierarchical CPL frontend.}
    \label{tab:chamfer_results}
    \begin{tabular}{lcc}
        \toprule
        Instance ID & RS & CPL Frontend  \\
        \midrule
         Airplane 655 & \textbf{0.01398} & 0.02975 \\
         Airplane 657 & \textbf{0.01707} & 0.02638 \\
         Airplane 719 & \textbf{0.01695} & 0.04460 \\
        \midrule
         Person 89  & \textbf{0.01373} & 0.02959 \\
         Person 91  & \textbf{0.01532} & 0.03200 \\
         Person 95  & \textbf{0.01151} & 0.02856 \\
        \bottomrule
    \end{tabular}
    \label{tab:chamfer_dist_rs_cpl}
\end{table}

\begin{itemize}
\item Local surrogate models like LIME-3D cannot evaluate every single point; they rely on clustering, producing explanations that are optimized for human perception.
\item CPL acts as an information funnel -- strictly optimized for classification. It isolates a minimum of points that are necessary to maximize classification confidence. What a neural network deems critical for classification must not align with human perception. 
\end{itemize}

Despite the stark difference in CF, CPL maintains operational utility. When testing our fused PointNet and CPL pipeline, where CPL was downsampled to 64 points, the network achieved an instance accuracy of $88.36\%$ across 5000 test instances. This performance is on par with our baseline metrics, where the network was trained on 64 random points ($89.01\%$). Thus, it introduces minimal or non-negative degradation to PointNet's classification performance, functioning as a capable edge accelerator. 
\section{Conclusion and Outlook}
\label{sec:conclusion}

\subsection{Conclusion}
We therefore showed the viability of deploying spatial deep learning architectures to edge processors. The contribution of our research is twofold, addressing the reality gap and algorithmic optimization.

First, we established a robust data-generation pipeline utilizing BLAINDER to generate realistic synthetic LiDAR data based on the ModelNet40 dataset. By emulating existing LiDAR sensors and introducing point-wise Gaussian noise, this pipeline simulates real-world sensor traits, such as view-dependent density. Cross-evaluation demonstrates that networks that are trained on pristine CAD models fail when subjected to sensor anomalies, but models that are trained on noisy data can abstract and achieve accurate results even on clean data. Furthermore, evaluations on variable densities reveal that PointNet is highly robust to data sparsity and maintains stable classification results even when aggressive data thinning is incorporated. 

Second, we implemented a pre-trained Critical Point Layer as a preprocessing filter.  Our results demonstrate that this frontend can deterministically compress a raw 1024-point cloud down to a sparse subset of 40–60 unique coordinates. When tested on the Raspberry Pi's CPU, this method bypasses the timings set by the FPS algorithm, proving that it is possible to run in real time on low-powered edge devices.  

\subsection{Outlook}

Several compelling avenues remain for optimization and architectural improvements: 
\begin{itemize}
    \item To scale this pipeline to handle millions of points simultaneously that are generated by an outdoor LiDAR scanner, future work can explore offloading the CPL shared-MLP onto a Field-Programmable Gate Array (FPGA) or specialized low-powered NPUs. FPGAs are well-suited for massive parallel execution, which would free up further CPU resources. 
    \item To resolve the extreme feature collapse where only a handful of points exist after the compression, it would be desirable to test continuous and fully differentiable operators, such as the Gumbel-Top-k approximation. Instead of hard filtering and row duplication to fill the rows, this optimization would compute an importance value for every point and could be sampled based on this score. 
\end{itemize}

%----------------------------------------------------------------------------------------
%	 REFERENCES
%----------------------------------------------------------------------------------------

\printbibliography % Output the bibliography

@article{li_lidar_2020,
	title = {Lidar for {Autonomous} {Driving}: {The} {Principles}, {Challenges}, and {Trends} for {Automotive} {Lidar} and {Perception} {Systems}},
	volume = {37},
	copyright = {https://ieeexplore.ieee.org/Xplorehelp/downloads/license-information/IEEE.html},
	issn = {1053-5888, 1558-0792},
	shorttitle = {Lidar for {Autonomous} {Driving}},
	url = {https://ieeexplore.ieee.org/document/9127855/},
	doi = {10.1109/MSP.2020.2973615},
	number = {4},
	urldate = {2026-03-02},
	journal = {IEEE Signal Processing Magazine},
	author = {Li, You and Ibanez-Guzman, Javier},
	month = jul,
	year = {2020},
	pages = {50--61},
}

@inproceedings{geiger_are_2012,
	title = {Are we ready for autonomous driving? {The} {KITTI} vision benchmark suite},
	issn = {1063-6919},
	shorttitle = {Are we ready for autonomous driving?},
	url = {https://ieeexplore.ieee.org/document/6248074},
	doi = {10.1109/CVPR.2012.6248074},
	urldate = {2026-03-02},
	booktitle = {2012 {IEEE} {Conference} on {Computer} {Vision} and {Pattern} {Recognition}},
	author = {Geiger, Andreas and Lenz, Philip and Urtasun, Raquel},
	month = jun,
	year = {2012},
	note = {ISSN: 1063-6919},
	pages = {3354--3361},
}

@article{geiger_vision_2013,
	title = {Vision meets robotics: {The} {KITTI} dataset},
	volume = {32},
	issn = {0278-3649},
	shorttitle = {Vision meets robotics},
	url = {https://doi.org/10.1177/0278364913491297},
	doi = {10.1177/0278364913491297},
	number = {11},
	urldate = {2026-03-03},
	journal = {Int. J. Rob. Res.},
	author = {Geiger, A and Lenz, P and Stiller, C and Urtasun, R},
	month = sep,
	year = {2013},
	pages = {1231--1237},
}

@misc{behley_semantickitti_2019,
	title = {{SemanticKITTI}: {A} {Dataset} for {Semantic} {Scene} {Understanding} of {LiDAR} {Sequences}},
	shorttitle = {{SemanticKITTI}},
	url = {http://arxiv.org/abs/1904.01416},
	doi = {10.48550/arXiv.1904.01416},
	urldate = {2026-03-04},
	publisher = {arXiv},
	author = {Behley, Jens and Garbade, Martin and Milioto, Andres and Quenzel, Jan and Behnke, Sven and Stachniss, Cyrill and Gall, Juergen},
	month = aug,
	year = {2019},
	note = {arXiv:1904.01416 [cs]},
}

@misc{wu_virtual_2023,
	title = {Virtual {Sparse} {Convolution} for {Multimodal} {3D} {Object} {Detection}},
	url = {http://arxiv.org/abs/2303.02314},
	doi = {10.48550/arXiv.2303.02314},
	urldate = {2026-03-05},
	publisher = {arXiv},
	author = {Wu, Hai and Wen, Chenglu and Shi, Shaoshuai and Li, Xin and Wang, Cheng},
	month = mar,
	year = {2023},
	note = {arXiv:2303.02314 [cs]},
}

@article{hu_learning_2022,
	title = {Learning {Semantic} {Segmentation} of {Large}-{Scale} {Point} {Clouds} {With} {Random} {Sampling}},
	volume = {44},
	issn = {1939-3539},
	url = {https://ieeexplore.ieee.org/abstract/document/9440696},
	doi = {10.1109/TPAMI.2021.3083288},
	number = {11},
	urldate = {2026-03-05},
	journal = {IEEE Transactions on Pattern Analysis and Machine Intelligence},
	author = {Hu, Qingyong and Yang, Bo and Xie, Linhai and Rosa, Stefano and Guo, Yulan and Wang, Zhihua and Trigoni, Niki and Markham, Andrew},
	month = nov,
	year = {2022},
	pages = {8338--8354},
}

@misc{qi_pointnet_2017,
	title = {{PointNet}: {Deep} {Learning} on {Point} {Sets} for {3D} {Classification} and {Segmentation}},
	shorttitle = {{PointNet}},
	url = {http://arxiv.org/abs/1612.00593},
	doi = {10.48550/arXiv.1612.00593},
	urldate = {2026-03-06},
	publisher = {arXiv},
	author = {Qi, Charles R. and Su, Hao and Mo, Kaichun and Guibas, Leonidas J.},
	month = apr,
	year = {2017},
	note = {arXiv:1612.00593 [cs]},
}

@misc{qi_pointnet_2017-1,
	title = {{PointNet}++: {Deep} {Hierarchical} {Feature} {Learning} on {Point} {Sets} in a {Metric} {Space}},
	shorttitle = {{PointNet}++},
	url = {http://arxiv.org/abs/1706.02413},
	doi = {10.48550/arXiv.1706.02413},
	urldate = {2026-03-06},
	publisher = {arXiv},
	author = {Qi, Charles R. and Yi, Li and Su, Hao and Guibas, Leonidas J.},
	month = jun,
	year = {2017},
	note = {arXiv:1706.02413 [cs]},
}

@misc{alonso_3d-mininet_2021,
	title = {{3D}-{MiniNet}: {Learning} a {2D} {Representation} from {Point} {Clouds} for {Fast} and {Efficient} {3D} {LIDAR} {Semantic} {Segmentation}},
	shorttitle = {{3D}-{MiniNet}},
	url = {http://arxiv.org/abs/2002.10893},
	doi = {10.48550/arXiv.2002.10893},
	urldate = {2026-03-09},
	publisher = {arXiv},
	author = {Alonso, Iñigo and Riazuelo, Luis and Montesano, Luis and Murillo, Ana C.},
	month = apr,
	year = {2021},
	note = {arXiv:2002.10893 [cs]},
}

@article{jia_ai-powered_2025,
	title = {{AI}-{Powered} {LiDAR} {Point} {Cloud} {Understanding} and {Processing}: {An} {Updated} {Survey}},
	volume = {26},
	issn = {1558-0016},
	shorttitle = {{AI}-{Powered} {LiDAR} {Point} {Cloud} {Understanding} and {Processing}},
	url = {https://ieeexplore.ieee.org/document/11021542/figures},
	doi = {10.1109/TITS.2025.3568500},
	number = {8},
	urldate = {2026-03-10},
	journal = {IEEE Transactions on Intelligent Transportation Systems},
	author = {Jia, Shanghui and Gong, Xinghan and Liu, Fang and Ma, Lingfei},
	month = aug,
	year = {2025},
	pages = {11249--11275},
}

@article{betsas_deep_2025,
	title = {Deep {Learning} on {3D} {Semantic} {Segmentation}: {A} {Detailed} {Review}},
	volume = {17},
	issn = {2072-4292},
	shorttitle = {Deep {Learning} on {3D} {Semantic} {Segmentation}},
	url = {http://arxiv.org/abs/2411.02104},
	doi = {10.3390/rs17020298},
	number = {2},
	urldate = {2026-03-11},
	journal = {Remote Sensing},
	author = {Betsas, Thodoris and Georgopoulos, Andreas and Doulamis, Anastasios and Grussenmeyer, Pierre},
	month = jan,
	year = {2025},
	note = {arXiv:2411.02104 [cs]},
	pages = {298},
}

@misc{wu_squeezesegv2_2018,
	title = {{SqueezeSegV2}: {Improved} {Model} {Structure} and {Unsupervised} {Domain} {Adaptation} for {Road}-{Object} {Segmentation} from a {LiDAR} {Point} {Cloud}},
	shorttitle = {{SqueezeSegV2}},
	url = {http://arxiv.org/abs/1809.08495},
	doi = {10.48550/arXiv.1809.08495},
	urldate = {2026-03-11},
	publisher = {arXiv},
	author = {Wu, Bichen and Zhou, Xuanyu and Zhao, Sicheng and Yue, Xiangyu and Keutzer, Kurt},
	month = sep,
	year = {2018},
	note = {arXiv:1809.08495 [cs]},
}

@misc{halperin_point_2025,
	title = {Point {Cloud} {Based} {Scene} {Segmentation}: {A} {Survey}},
	shorttitle = {Point {Cloud} {Based} {Scene} {Segmentation}},
	url = {http://arxiv.org/abs/2503.12595},
	doi = {10.48550/arXiv.2503.12595},
	urldate = {2026-03-13},
	publisher = {arXiv},
	author = {Halperin, Dan and Eisl, Niklas},
	month = mar,
	year = {2025},
	note = {arXiv:2503.12595 [cs]},
}

@inproceedings{tang_searching_2020,
	address = {Cham},
	title = {Searching {Efficient} {3D} {Architectures} with {Sparse} {Point}-{Voxel} {Convolution}},
	isbn = {978-3-030-58604-1},
	doi = {10.1007/978-3-030-58604-1_41},
	language = {en},
	booktitle = {Computer {Vision} – {ECCV} 2020},
	publisher = {Springer International Publishing},
	author = {Tang, Haotian and Liu, Zhijian and Zhao, Shengyu and Lin, Yujun and Lin, Ji and Wang, Hanrui and Han, Song},
	editor = {Vedaldi, Andrea and Bischof, Horst and Brox, Thomas and Frahm, Jan-Michael},
	year = {2020},
	pages = {685--702},
}

@article{wisultschew_characterizing_2022,
	title = {Characterizing {Deep} {Neural} {Networks} on {Edge} {Computing} {Systems} for {Object} {Classification} in {3D} {Point} {Clouds}},
	volume = {22},
	issn = {1558-1748},
	url = {https://ieeexplore.ieee.org/abstract/document/9843856},
	doi = {10.1109/JSEN.2022.3193060},
	number = {17},
	urldate = {2026-03-17},
	journal = {IEEE Sensors Journal},
	author = {Wisultschew, Cristian and Pérez, Alejandro and Otero, Andrés and Mujica, Gabriel and Portilla, Jorge},
	month = sep,
	year = {2022},
	pages = {17075--17089},
}

@misc{wu_3d_2015,
	title = {{3D} {ShapeNets}: {A} {Deep} {Representation} for {Volumetric} {Shapes}},
	shorttitle = {{3D} {ShapeNets}},
	url = {http://arxiv.org/abs/1406.5670},
	doi = {10.48550/arXiv.1406.5670},
	urldate = {2026-03-19},
	publisher = {arXiv},
	author = {Wu, Zhirong and Song, Shuran and Khosla, Aditya and Yu, Fisher and Zhang, Linguang and Tang, Xiaoou and Xiao, Jianxiong},
	month = apr,
	year = {2015},
	note = {arXiv:1406.5670 [cs]},
}

@article{reitmann_blainderblender_2021-1,
	title = {{BLAINDER}—{A} {Blender} {AI} {Add}-{On} for {Generation} of {Semantically} {Labeled} {Depth}-{Sensing} {Data}},
	volume = {21},
	copyright = {http://creativecommons.org/licenses/by/3.0/},
	issn = {1424-8220},
	url = {https://www.mdpi.com/1424-8220/21/6/2144},
	doi = {10.3390/s21062144},
	language = {en},
	number = {6},
	urldate = {2026-04-01},
	journal = {Sensors},
	publisher = {Multidisciplinary Digital Publishing Institute},
	author = {Reitmann, Stefan and Neumann, Lorenzo and Jung, Bernhard},
	month = jan,
	year = {2021},
	pages = {2144},
}

@misc{saeid_enhancing_2025,
	title = {Enhancing {3D} {Point} {Cloud} {Classification} with {ModelNet}-{R} and {Point}-{SkipNet}},
	url = {http://arxiv.org/abs/2509.05198},
	doi = {10.48550/arXiv.2509.05198},
	urldate = {2026-04-02},
	publisher = {arXiv},
	author = {Saeid, Mohammad and Salarpour, Amir and MohajerAnsari, Pedram},
	month = sep,
	year = {2025},
	note = {arXiv:2509.05198 [cs]},
}

@misc{mohammadi_point-gn_2024,
	title = {Point-{GN}: {A} {Non}-{Parametric} {Network} {Using} {Gaussian} {Positional} {Encoding} for {Point} {Cloud} {Classification}},
	shorttitle = {Point-{GN}},
	url = {http://arxiv.org/abs/2412.03056},
	doi = {10.48550/arXiv.2412.03056},
	urldate = {2026-04-07},
	publisher = {arXiv},
	author = {Mohammadi, Marzieh and Salarpour, Amir},
	month = dec,
	year = {2024},
	note = {arXiv:2412.03056 [cs]},
}

@article{zhu_advancements_2024,
	title = {Advancements in {Point} {Cloud} {Data} {Augmentation} for {Deep} {Learning}: {A} {Survey}},
	volume = {153},
	issn = {00313203},
	shorttitle = {Advancements in {Point} {Cloud} {Data} {Augmentation} for {Deep} {Learning}},
	url = {http://arxiv.org/abs/2308.12113},
	doi = {10.1016/j.patcog.2024.110532},
	urldate = {2026-04-13},
	journal = {Pattern Recognition},
	author = {Zhu, Qinfeng and Fan, Lei and Weng, Ningxin},
	month = sep,
	year = {2024},
	note = {arXiv:2308.12113 [cs]},
	pages = {110532},
}

@article{van_den_herrewegen_point_2023,
	title = {Point {Cloud} {Classification} with {ModelNet40}: {What} is left?},
	language = {en},
	author = {Van den Herrewegen, Jarne and Tourwé, Tom and Wyffels, Francis},
	year = {2023},
}

@misc{sun_benchmarking_2022,
	title = {Benchmarking {Robustness} of {3D} {Point} {Cloud} {Recognition} {Against} {Common} {Corruptions}},
	url = {http://arxiv.org/abs/2201.12296},
	doi = {10.48550/arXiv.2201.12296},
	urldate = {2026-04-28},
	publisher = {arXiv},
	author = {Sun, Jiachen and Zhang, Qingzhao and Kailkhura, Bhavya and Yu, Zhiding and Xiao, Chaowei and Mao, Z. Morley},
	month = jan,
	year = {2022},
	note = {arXiv:2201.12296 [cs]},
}

@misc{levi_fast_2024,
	title = {Fast and {Simple} {Explainability} for {Point} {Cloud} {Networks}},
	url = {http://arxiv.org/abs/2403.07706},
	doi = {10.48550/arXiv.2403.07706},
	urldate = {2026-04-29},
	publisher = {arXiv},
	author = {Levi, Meir Yossef and Gilboa, Guy},
	month = mar,
	year = {2024},
	note = {arXiv:2403.07706 [cs]},
}

@misc{zheng_pointcloud_2019,
	title = {{PointCloud} {Saliency} {Maps}},
	url = {http://arxiv.org/abs/1812.01687},
	doi = {10.48550/arXiv.1812.01687},
	urldate = {2026-04-29},
	publisher = {arXiv},
	author = {Zheng, Tianhang and Chen, Changyou and Yuan, Junsong and Li, Bo and Ren, Kui},
	month = sep,
	year = {2019},
	note = {arXiv:1812.01687 [cs]},
}

@misc{tan_surrogate_2021,
	title = {Surrogate {Model}-{Based} {Explainability} {Methods} for {Point} {Cloud} {NNs}},
	url = {http://arxiv.org/abs/2107.13459},
	doi = {10.48550/arXiv.2107.13459},
	urldate = {2026-04-29},
	publisher = {arXiv},
	author = {Tan, Hanxiao and Kotthaus, Helena},
	month = aug,
	year = {2021},
	note = {arXiv:2107.13459 [cs]},
}

@inproceedings{tan_visualizing_2023,
	title = {Visualizing {Global} {Explanations} of {Point} {Cloud} {DNNs}},
	issn = {2642-9381},
	url = {https://ieeexplore.ieee.org/document/10030393},
	doi = {10.1109/WACV56688.2023.00472},
	urldate = {2026-05-11},
	booktitle = {2023 {IEEE}/{CVF} {Winter} {Conference} on {Applications} of {Computer} {Vision} ({WACV})},
	author = {Tan, Hanxiao},
	month = jan,
	year = {2023},
	note = {ISSN: 2642-9381},
	pages = {4730--4739},
}

@article{zhang_deep_2023,
	title = {Deep {Learning}-based {3D} {Point} {Cloud} {Classification}: {A} {Systematic} {Survey} and {Outlook}},
	volume = {79},
	issn = {01419382},
	shorttitle = {Deep {Learning}-based {3D} {Point} {Cloud} {Classification}},
	url = {http://arxiv.org/abs/2311.02608},
	doi = {10.1016/j.displa.2023.102456},
	urldate = {2026-05-18},
	journal = {Displays},
	author = {Zhang, Huang and Wang, Changshuo and Tian, Shengwei and Lu, Baoli and Zhang, Liping and Ning, Xin and Bai, Xiao},
	month = sep,
	year = {2023},
	note = {arXiv:2311.02608 [cs.CV]},
	pages = {102456},
}

@misc{nezhadarya_adaptive_2020,
	title = {Adaptive {Hierarchical} {Down}-{Sampling} for {Point} {Cloud} {Classification}},
	url = {http://arxiv.org/abs/1904.08506},
	doi = {10.48550/arXiv.1904.08506},
	urldate = {2026-05-18},
	publisher = {arXiv},
	author = {Nezhadarya, Ehsan and Taghavi, Ehsan and Razani, Ryan and Liu, Bingbing and Luo, Jun},
	month = may,
	year = {2020},
	note = {arXiv:1904.08506 [cs.CV]},
}

@misc{bhatia_survey_2010,
	title = {Survey of {Nearest} {Neighbor} {Techniques}},
	url = {http://arxiv.org/abs/1007.0085},
	doi = {10.48550/arXiv.1007.0085},
	urldate = {2026-06-08},
	publisher = {arXiv},
	author = {Bhatia, Nitin and Vandana},
	month = jul,
	year = {2010},
	note = {arXiv:1007.0085 [cs.CV]},
}

@misc{wu_squeezeseg_2017,
	title = {{SqueezeSeg}: {Convolutional} {Neural} {Nets} with {Recurrent} {CRF} for {Real}-{Time} {Road}-{Object} {Segmentation} from {3D} {LiDAR} {Point} {Cloud}},
	shorttitle = {{SqueezeSeg}},
	url = {http://arxiv.org/abs/1710.07368},
	doi = {10.48550/arXiv.1710.07368},
	urldate = {2026-06-08},
	publisher = {arXiv},
	author = {Wu, Bichen and Wan, Alvin and Yue, Xiangyu and Keutzer, Kurt},
	month = oct,
	year = {2017},
	note = {arXiv:1710.07368 [cs.CV]},
}

@misc{wang_dynamic_2019,
	title = {Dynamic {Graph} {CNN} for {Learning} on {Point} {Clouds}},
	url = {http://arxiv.org/abs/1801.07829},
	doi = {10.48550/arXiv.1801.07829},
	urldate = {2026-06-09},
	publisher = {arXiv},
	author = {Wang, Yue and Sun, Yongbin and Liu, Ziwei and Sarma, Sanjay E. and Bronstein, Michael M. and Solomon, Justin M.},
	month = jun,
	year = {2019},
	note = {arXiv:1801.07829 [cs.CV]},
}

@misc{knaebel_point2vec_2023,
	title = {{Point2Vec} for {Self}-{Supervised} {Representation} {Learning} on {Point} {Clouds}},
	url = {http://arxiv.org/abs/2303.16570},
	doi = {10.48550/arXiv.2303.16570},
	urldate = {2026-06-09},
	publisher = {arXiv},
	author = {Knaebel, Karim and Schult, Jonas and Hermans, Alexander and Leibe, Bastian},
	month = oct,
	year = {2023},
	note = {arXiv:2303.16570 [cs.CV]},
}

@misc{zhang_pointcutmix_2021,
	title = {{PointCutMix}: {Regularization} {Strategy} for {Point} {Cloud} {Classification}},
	shorttitle = {{PointCutMix}},
	url = {http://arxiv.org/abs/2101.01461},
	doi = {10.48550/arXiv.2101.01461},
	urldate = {2026-06-10},
	publisher = {arXiv},
	author = {Zhang, Jinlai and Chen, Lyujie and Ouyang, Bo and Liu, Binbin and Zhu, Jihong and Chen, Yujing and Meng, Yanmei and Wu, Danfeng},
	month = feb,
	year = {2021},
	note = {arXiv:2101.01461 [cs.CV]},
}

@inproceedings{wu_point_2024,
	title = {Point {Transformer} {V3}: {Simpler}, {Faster}, {Stronger}},
	issn = {2575-7075},
	shorttitle = {Point {Transformer} {V3}},
	url = {https://ieeexplore.ieee.org/document/10658198/},
	doi = {10.1109/CVPR52733.2024.00463},
	urldate = {2026-06-22},
	booktitle = {2024 {IEEE}/{CVF} {Conference} on {Computer} {Vision} and {Pattern} {Recognition} ({CVPR})},
	author = {Wu, Xiaoyang and Jiang, Li and Wang, Peng-Shuai and Liu, Zhijian and Liu, Xihui and Qiao, Yu and Ouyang, Wanli and He, Tong and Zhao, Hengshuang},
	month = jun,
	year = {2024},
	note = {ISSN: 2575-7075},
	pages = {4840--4851},
}

@misc{Meyer2026SyntheticLiDAR,
  author       = {Meyer, N. and Reitmann, S.},
  title        = {Synthetic LiDAR Data Generation and Deterministic Downsampling for Point Cloud Classification on the Edge},
  year         = {2026},
  month        = aug,
  publisher    = {Zenodo},
  doi          = {10.5281/zenodo.21835460},
  url          = {https://doi.org/10.5281/zenodo.21835460}
}
\clearpage

\appendix
\section{Appendix}
% Force the layout to span both columns for the Appendix header and image
\begin{minipage}{\textwidth}
    %\section{Appendix}
    %\vspace{0.2cm} % Add some breathing room below the title
    
    \centering
    % Scaled slightly to 0.82 to guarantee the header, giant matrix, and caption all fit on this single sheet
    \includegraphics[angle=90, width=0.92\textwidth, height=0.92\textheight, keepaspectratio]{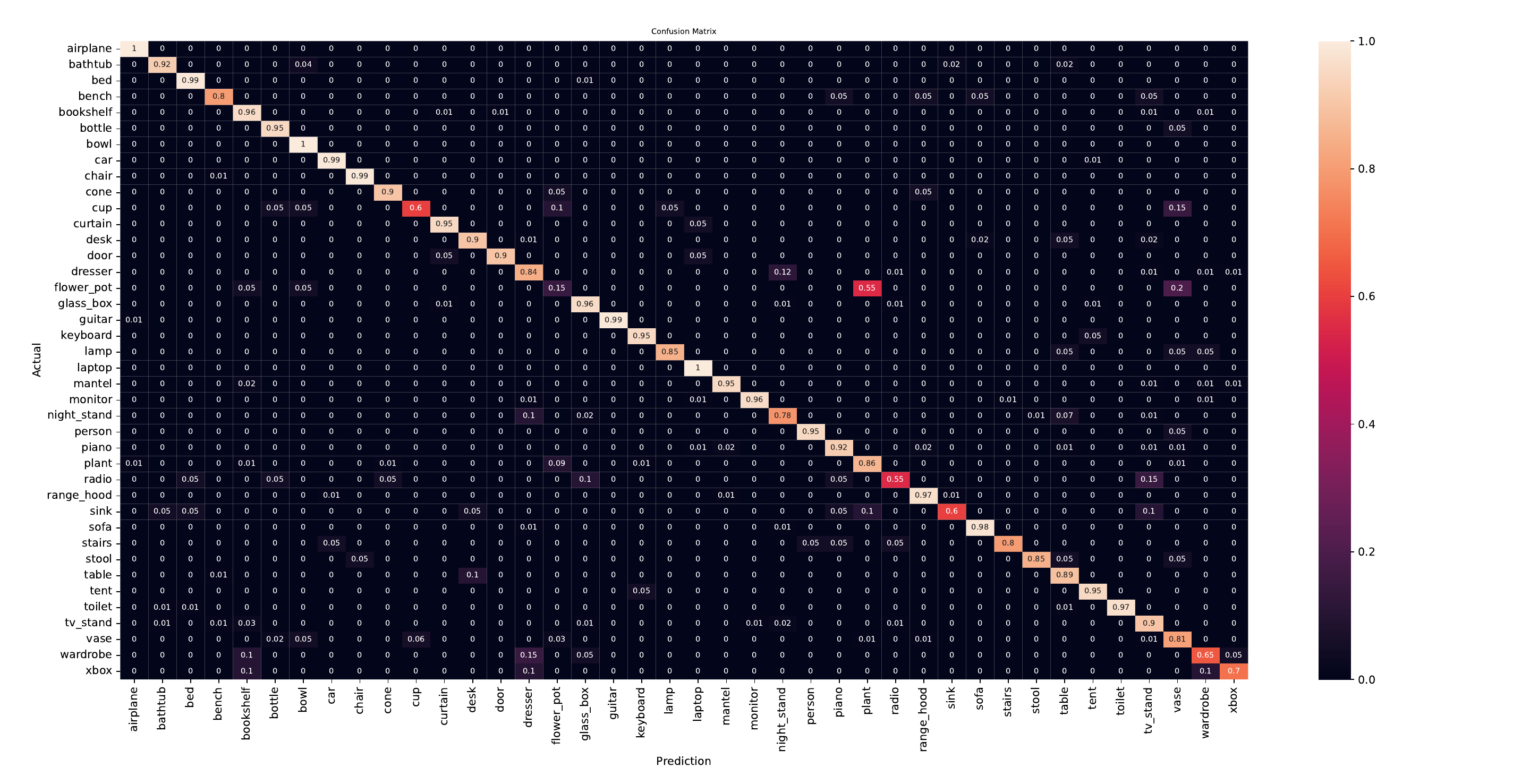}
    
    %\vspace{0.3cm}
    % Captions the image perfectly across both columns
    \captionof{figure}{Confusion Matrix displaying the results of PoinNet evaluated on ModelNet. It can be seen that the network has difficulty differentiating between several classes, such as 'flower pot' and 'plant'.}
    \label{fig:confusion_matrix_pointnet_modelnet}
\end{minipage}

%----------------------------------------------------------------------------------------

\end{document}